\documentclass[lettersize,journal]{IEEEtran}

\usepackage{amsmath,amsfonts}
\usepackage{algorithmic}
\usepackage{algorithm}
\usepackage{array}
\usepackage[caption=false,font=normalsize,labelfont=sf,textfont=sf]{subfig}
\usepackage{textcomp}
\usepackage{stfloats}
\usepackage{url}
\usepackage{verbatim}
\usepackage{graphicx}
\usepackage{cite}
\usepackage{booktabs}
\usepackage{multirow}
\usepackage{array}
\usepackage{amsmath}
\usepackage{amsfonts}
\usepackage{xcolor}
\usepackage{stackengine}
\usepackage{makecell}
\usepackage[most]{tcolorbox}
\usepackage{mdframed}
\usepackage{longtable}
\usepackage{tabularx}
\usepackage{adjustbox}
\usepackage{fontawesome}
\usepackage{cuted}
\usepackage{placeins}
\usepackage{float}

\newcolumntype{C}{>{\centering\arraybackslash}X}

\usepackage[dvipsnames]{xcolor}
\usepackage{graphicx}
\usepackage{subcaption}
\usepackage{booktabs}
\usepackage{amssymb}
\usepackage{colortbl}
\usepackage[
  pagebackref=true,
  breaklinks=true,
  bookmarks=true,
  colorlinks,
  citecolor=blue,
  linkcolor=blue,
  urlcolor=blue
]{hyperref}

\usepackage{pgfplots}
\pgfplotsset{compat=1.18}

\usepackage{tabularx}
\usepackage{booktabs}
\usepackage{ragged2e}
\newcolumntype{L}{>{\RaggedRight\arraybackslash}X}
\patchcmd{\IEEEbiography}{8pt}{0pt}{}{}

\begin{document}

\title{Lift3D-VLA: Lifting VLA Models to 3D Geometry and Dynamics-Aware Manipulation}

\author{Jiaming Liu$^\dagger$,~\IEEEmembership{Student Member,~IEEE,} Qingpo Wuwu$^\dagger$, Nuowei Han$^\dagger$, Hao Chen$^\dagger$, Zhuoyang Liu, Fan~Fei, \\Yueru Jia, Chenyang Gu, Yandong Guo, Boxin Shi,~\IEEEmembership{Senior Member,~IEEE,}
and Shanghang Zhang%

\thanks{Jiaming Liu, Qingpo Wuwu, Nuowei Han, Zhuoyang Liu, Yueru Jia, Chenyang Gu, Fan Fei, Boxin Shi, and Shanghang Zhang are with the State Key Laboratory of Multimedia Information Processing and National Engineering Research Center of Visual Technology, School of Computer Science, Peking University, Beijing, China. Hao Chen is with the CUHK, Shatin, Hong Kong. Yandong Guo is with the AI$^{2}$Robotics, Beijing, China.}
\thanks{
$^\dagger$Jiaming Liu, Qingpo Wuwu, Nuowei Han, and Hao Chen contributed equally as co-first authors. Corresponding author: Shanghang Zhang. E-mail: \{jiamingliu@stu.pku.edu.cn, shanghang@pku.edu.cn\}}
}

\markboth{}{}


\maketitle
\thispagestyle{plain}
\pagestyle{plain}

\begin{abstract}
Recently, Vision–Language–Action (VLA) models have demonstrated strong generalization across diverse tasks. However, effective robotic manipulation in physical environments fundamentally requires geometric understanding and  spatial reasoning. While some VLA approaches attempt to incorporate 3D information, they are constrained by limited data availability and geometric information loss in current 3D encoding pipelines, and fail to jointly capture 3D geometry and temporally structured actions in dynamic environments.
To address these limitations, we introduce Lift3D-VLA, a unified VLA framework that equips models with explicit 3D point cloud reasoning and enables temporally coherent action generation.
First, building upon our previous work Lift3D, an enhanced 2D model-lifting strategy is proposed to geometrically align 3D points with pretrained 2D positional embeddings. This design enables direct point-cloud encoding within the VLA vision encoder while minimizing spatial information loss.
Based on explicit 3D inputs, we propose Geometry-Centric Masked Autoencoding (GC-MAE), a dual-objective self-supervised framework that reconstructs the current point cloud while predicting its future geometric evolution. This formulation allows the 2D vision encoder to internalize both 3D structure and physical dynamics.
To fully exploit 3D representations, we further design layer-wise temporal action modeling, which leverages multiple layers of the LLM to collaboratively predict action chunks, enabling temporally consistent predictions.
Across 22 simulated tasks and 8 real-world manipulation tasks, Lift3D-VLA achieves 10.8\% and 11.1\% higher mean success rates on MetaWorld and RLBench than the best-performing prior VLA methods, and outperforms the strongest real-world baseline by 4 percentage points, while exhibiting stronger generalization to out-of-distribution perturbations.
Project website: \url{https://lift3dvla.github.io/}.
\end{abstract}


\begin{figure*}[!t]
\includegraphics[width=0.99\textwidth]{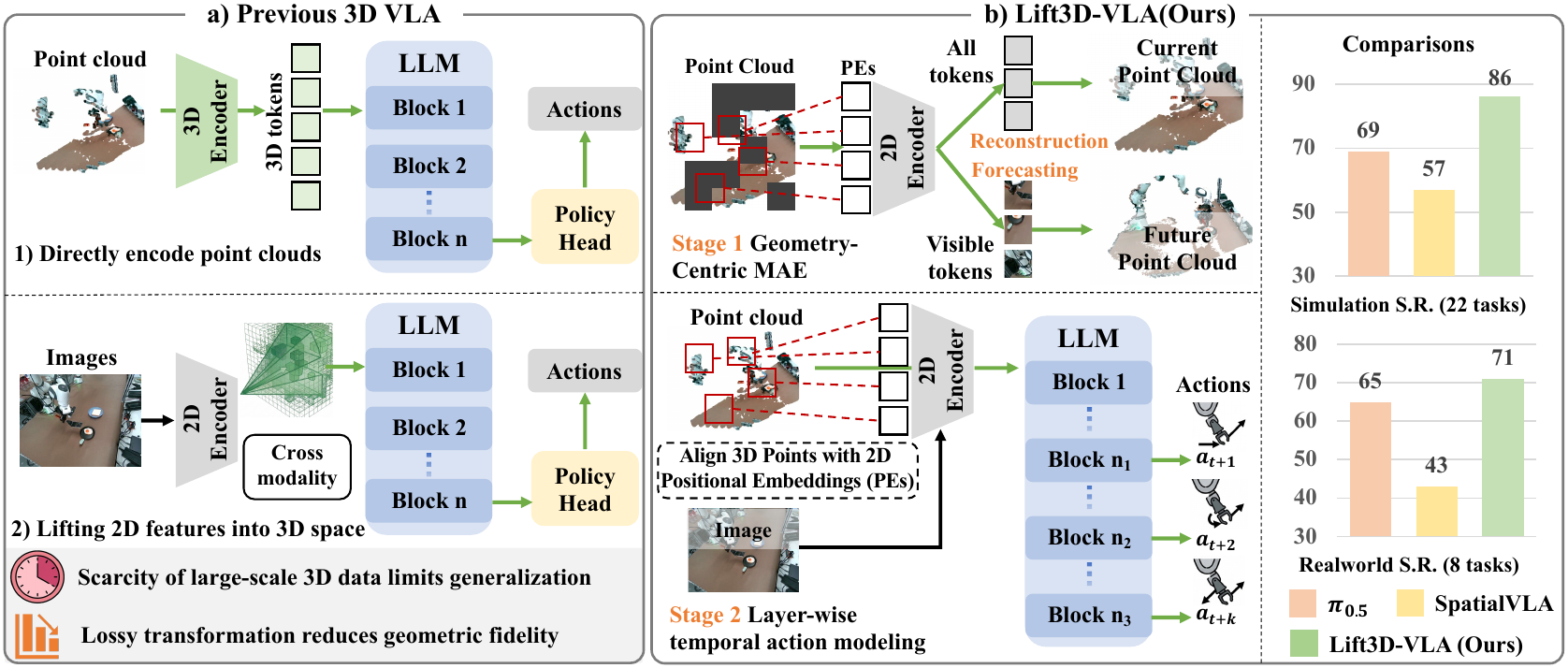}
    \captionof{figure}{\textbf{Overview.}
a) Unlike previous 3D VLA methods that encode point clouds either with newly introduced 3D encoders or by projecting features between 2D and 3D spaces, b) We propose \textbf{Lift3D-VLA}, equipping 2D VLA models with explicit 3D reasoning and temporally coherent action generation.
First, following our prior work Lift3D~\cite{jia2025lift3d}, we align 3D points with 2D positional embeddings to enable direct point-cloud encoding. Building on this, in Stage 1 we introduce Geometry-Centric MAE, which reconstructs the present point cloud while predicting its future geometric evolution, enhancing 3D representations. In Stage 2, we further propose layer-wise temporal action modeling, leveraging multiple layers of the LLM to produce temporally consistent actions. Across 22 simulated and 8 real-world tasks, Lift3D-VLA achieves SOTA performance.
}
\label{fig:teaser}
\vspace{-0.1cm}
\end{figure*}

\section{Introduction}
Vision–Language–Action (VLA) models have recently emerged as a promising paradigm for robotic manipulation~\cite{zitkovich2023rt, 2025_3_13_HybridVLA,2025_4_22_pi0_5}, showing strong generalization across diverse tasks by combining visual perception, language conditioning, and action generation. Despite this progress, robotic manipulation fundamentally requires spatial reasoning in the physical world~\cite{shridhar2022cliport, zhu2024point, eisner2022flowbot3d, fang2023anygrasp, shridhar2023perceiver}: the robot must infer 3D structure, reason about geometric relationships (e.g., reachability, occlusion, and contact), and plan actions that remain temporally consistent as the geometry evolves. Purely 2D VLA pipelines often struggle to reliably capture these geometric constraints, particularly in cluttered or dynamic environments.

A natural direction is to explicitly inject 3D information into VLA models or manipulation policies, with existing approaches primarily falling into two paradigms, as shown in Figure~\ref{fig:teaser} a).
First, some methods directly encode point clouds, voxels, or multi-view observations~\cite{shridhar2023perceiver, chen2023polarnet, liu2022frame, zhang2024leveraging, wang2024rise, ze20243d, james2022coarse, li2026pointvla}. However, unlike large-scale 2D vision–language pretraining, large robotic 3D datasets and strong 3D foundation encoders remain scarce, making it difficult to learn transferable and generalizable 3D representations.
Second, other approaches rely on cross-modal transformations, such as lifting 2D features into 3D space~\cite{gervet2023act3d, ke20243d, xian2023chaineddiffuser, shridhar2022cliport, qu2025spatialvla} or projecting 3D point clouds into multi-view images~\cite{goyal2023rvt, wang2024vihe, zhang2024sam, 2025_8_08_3DS_VLA}. However, such transformations are inherently lossy, compromising geometric fidelity and weakening the structural correspondence between 2D pretrained representations and 3D spatial structure. This ultimately limits scalability and diminishes the benefits of large-scale 2D pretraining.

To mitigate these bottlenecks, our prior work Lift3D~\cite{jia2025lift3d} innovatively proposes a 2D-pretraining reuse paradigm to endow 2D foundation models with 3D geometric perception for robotic manipulation. Rather than training a new 3D foundation model from scratch, Lift3D progressively augments a pretrained 2D backbone with implicit 3D representation enhancement via a task-aware masked autoencoder (MAE), and an explicit point-cloud encoding that aligns 3D points with pretrained 2D positional priors.
This design largely eliminates the need for massive 3D pretraining data, reduces spatial fidelity loss introduced by cross-modal transformations, and maintains computational efficiency and scalability.

Despite its effectiveness, Lift3D still exhibits two limitations that become critical when applied to VLA models in dynamic environments. \textbf{First}, its implicit enhancement remains indirect: the masked reconstruction objective operates primarily on RGB and depth signals, which do not explicitly enforce learning of coherent 3D structure. As a result, the model struggles to acquire robust geometric understanding that generalizes across viewpoints and physical interactions.
\textbf{Second}, Lift3D is not designed to jointly model evolving 3D geometry and temporally structured actions, which can lead to brittle behavior when long-horizon, time-consistent decisions are required. In dynamic manipulation scenarios, an agent must reason about how the scene geometry evolves over time while generating temporally coherent action sequences~\cite{chenac, fu2024mobile, huang2025thinkact}. 
These limitations highlight the need for a framework that learns rich 3D structure, models how geometry evolves, and generates temporally consistent actions.

In this paper, as shown in Figure~\ref{fig:teaser} b), we propose Lift3D-VLA, a unified VLA framework that enables explicit 3D point-cloud reasoning and temporally structured action generation.
Building upon the core philosophy of Lift3D, we inherit the 2D model lifting strategy to enable faithful and efficient explicit 3D encoding. Specifically, we project point cloud data onto multiple virtual planes, establishing a structured alignment between 3D geometry and 2D positional embeddings. To avoid geometric distortion, we further leverage camera parameters to anchor the front virtual plane to the camera viewpoint. This simplified design enables point-cloud tokens to be processed by a pretrained 2D VLA model while minimizing spatial information loss.
To improve the 3D representations learned by the VLA vision encoder, we propose Geometry-Centric Masked Autoencoding (GC-MAE). Given explicit 3D input, GC-MAE reconstructs the corresponding point cloud under a temporally asymmetric self-supervision scheme, jointly recovering present geometry and predicting its future geometric evolution. This recovery–forecast formulation encourages the model to capture physical dynamics in addition to static structure, enabling the VLA’s 2D encoder to develop a robust understanding of 3D physical properties.
Building upon these enriched 3D representations, we introduce layer-wise temporal action modeling to improve action generation in dynamic environments. Instead of relying on a separate decoding head to predict action chunks~\cite{2025_4_22_pi0_5, wen2024diffusion, bi2025motus}, our method leverages sequence representations from intermediate to deep LLM layers to capture temporal dependencies. Multiple layers consecutively predict sequential steps within the action chunk, resulting in temporally coherent and execution-stable actions.

Compared with Lift3D, we scale up pretraining to equip Lift3D-VLA with richer pretraining knowledge. Specifically, we perform GC-MAE self-supervised training on 140K trajectories and robotic pretraining on 400K trajectories collected from large-scale robotic datasets~\cite{open_x_embodiment_rt_x_2023, khazatsky2024droid, wu2024robomind}. The downstream evaluation is also extended by introducing dual-arm tasks beyond those considered in Lift3D. Across 22 simulated tasks and 8 real-world tasks, Lift3D-VLA achieves 10.8\% and 11.1\% higher mean success rates on MetaWorld \cite{2019_10_24_Meta_World} and RLBench \cite{2019_9_26_RLBench} than prior state-of-the-art VLA methods, respectively, and outperforms the strongest real-world baseline.
We further validate its long-horizon manipulation capability on tasks such as repeatedly scooping eggs from a pan under continuously changing conditions, while demonstrating strong generalization to unseen objects, backgrounds, and lighting variations.
In summary, our main contributions are:

\begin{itemize}

\item We extend our preliminary work~\cite{jia2025lift3d} into a unified Lift3D-VLA framework that integrates explicit 3D point-cloud reasoning with temporally structured action generation, supported by large-scale robotic pretraining. The key improvements include:

\item We propose Geometry-Centric Masked Autoencoding, a self-supervised scheme that reconstructs the present point cloud while forecasting its future geometric evolution, enabling the VLA's vision encoder to learn more robust 3D representations.

\item We develop a layer-wise temporal action modeling strategy that leverages sequence representations from consecutive LLM layers to predict action steps, improving temporal coherence in dynamic physical environments.

\end{itemize}

\section{Related Work}
\subsection{Representation Learning for Robotics}
Recent progress in pretrained visual representations has been largely driven by self-supervised learning paradigms such as contrastive learning~\cite{2020_7_21_PointContrast}, self-distillation~\cite{caron2020unsupervised}, and MAE~\cite{he2022masked}. Building upon these advances, several works aim to enhance visual representations to better support robotic perception and control. For instance, R3M~\cite{2022_11_18_R3M} learns universal embodied representations from large-scale human video data using contrastive learning, VIP~\cite{li2025vipvisioninstructedpretraining} produces dense reward functions for unseen robotic tasks, and MVP~\cite{2022_3_11_MVP}, VC-1~\cite{2023_3_31_VC_1}, and Voltron~\cite{karamcheti2023languagedrivenrepresentationlearningrobotics} explore MAE-style pretraining for robotic perception. However, these approaches are still primarily built upon 2D visual representations, which often lack the spatial fidelity required for complex manipulation.
To address this limitation, recent work explores 3D-aware robotic representations. One approach learns geometry-aware representations through multi-view modeling or reconstruction, where methods such as MV-MWM~\cite{seo2023multi}, 3D-MVP~\cite{qian20253d}, SPA~\cite{zhu2024spa}, CL3R~\cite{cui2025cl3r}, and HyperMVP~\cite{yang2026hyperbolic} learn consistent geometric features across views and align 3D structure with pretrained 2D semantic representations. 
Another approaches~\cite{yang2026robo3r, zhang2025canonical} integrate geometric modeling more directly into policy learning frameworks, which improve spatial grounding through multi-view geometry, robot-centric reconstruction, canonical 3D coordinates, or fused RGB-D scene representations. 
Different from the aforementioned approaches, our previous work Lift3D~\cite{jia2025lift3d} leverages large-scale 2D foundation models and progressively augments them with implicit 3D representations, enabling enhanced spatial modeling while avoiding learning pretrained knowledge from scratch.
Building upon Lift3D, we further propose Geometry-Centric Masked Autoencoding, a dual-objective self-supervised framework that reconstructs the present point cloud while predicting its future state, which explicitly enforces the learning of coherent 3D geometric structure and physical dynamics.

\subsection{Vision-Language-Action Models}
VLA models extend pretrained vision-language models (VLMs) to robotic control, enabling robots to follow natural language instructions while benefiting from large-scale semantic priors~\cite{2024_2_12_Prismatic_VLMs, 2024_9_05_OpenVLA}. Early VLA research primarily focused on scaling robot demonstration data and adapting pretrained  VLMs for policy learning~\cite{2024_9_05_OpenVLA}, demonstrating improved generalization across manipulation tasks. Subsequent works further improve the expressiveness of VLA policies by introducing continuous generative policy heads, where diffusion-based approaches model complex action distributions through iterative denoising~\cite{wen2024diffusion, 2025_3_13_HybridVLA}, while flow-matching formulations provide a more efficient alternative for continuous action generation~\cite{2024_10_31_pi0, 2025_4_22_pi0_5}.
Building upon these generation paradigms, another line of VLA research~\cite{zhao2025cot,zhang2025upvlaunifiedunderstandingprediction} investigates the use of future state prediction to facilitate physical world modeling and improve action generation.
Motivated by the need for richer and more unified representations, recent approaches attempt to integrate generative and understanding capabilities within a single model using mixture-of-transformers architectures~\cite{bi2025motus, kim2026cosmos, gu2025manualvla, liu2026last, chen2026mv, liu2026last_hd}.
Beyond 2D representations, some works focus on enhancing spatial understanding and reasoning in VLA models. Methods such as 3DS-VLA~\cite{2025_8_08_3DS_VLA}, SpatialVLA~\cite{qu2025spatialvla}, and PointVLA~\cite{li2026pointvla} introduce geometric information by aligning visual features with spatial coordinates or incorporating point cloud observations, while other approaches extend the perception space toward multisensory representations by integrating visual, depth, or tactile signals~\cite{liu2025mla}.
In contrast, Lift3D-VLA builds upon Lift3D’s 2D model lifting strategy, allowing for the explicit incorporation of 3D point clouds, and further introduces layer-wise temporal action modeling to generate temporally coherent actions from 3D inputs.

\section{Preliminaries}
\label{sec/preliminaries}

\begin{figure*}[t]
    \centering
    \includegraphics[width=\textwidth]{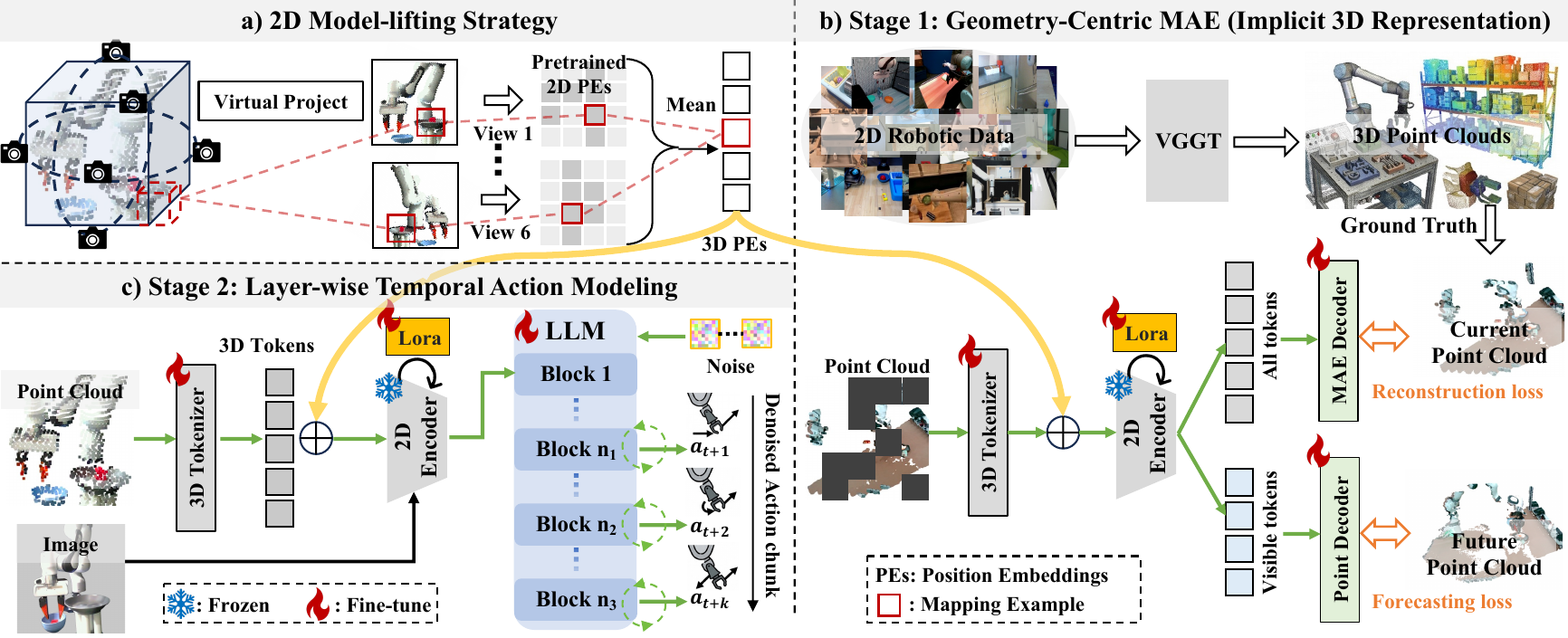}
    \caption{\textbf{Lift3D-VLA Framework.} \textbf{a)} Following our previous work Lift3D, we perform virtual projection to align 3D points with pretrained 2D positional embeddings (PEs), thereby constructing geometry-aligned 3D PEs that enable the 2D vision encoder in VLA models to directly process point cloud inputs. \textbf{b)} Stage 1. To enhance 3D physical representations, we first leverage VGGT to synthesize 3D point clouds from large-scale robotic data, which serve as self-supervised training targets. We then introduce Geometry-Centric MAE, a framework that reconstructs the current point cloud while simultaneously predicting its future geometric evolution, enabling the model to capture physical dynamics in addition to static spatial structure. \textbf{c)} Stage 2. Building on the strengthened 3D representations, we further propose layer-wise temporal action modeling, which leverages the sequence modeling capabilities of intermediate and deep LLM layers to generate temporally consistent action sequences.
    }
    \label{fig:1_main_pipeline}
\end{figure*}

\textbf{Task Formulation.}
VLA models aim to enable robots to execute diverse tasks by mapping visual observations and language instructions to action sequences. Formally, given a natural language task instruction $\ell$ and observation $o_t$ at timestep $t$, a VLA policy $\pi_\theta$ predicts an action chunk $(a_t, a_{t+1}, \ldots, a_{t+H-1})$ for task execution:
\begin{equation}
    \pi_{\theta}: (\ell, o_t) \rightarrow (a_t, a_{t+1}, \ldots, a_{t+H-1}),
\end{equation}
where $H$ denotes the action chunk size and the observation $o_t = \{I^1_t, \ldots, I^n_t, P_t, q_t\}$ comprises $n$ camera images, point cloud $P_t$, and proprioceptive state $q_t$ (e.g., joint positions). Each action $a_t \in \mathbb{R}^7$ represents the end-effector pose:
\begin{equation}
    a_t = [\Delta x, \Delta y, \Delta z, R_r, R_p, R_y, g],
\end{equation}
where $\Delta x, \Delta y, \Delta z$ represent relative Cartesian position offsets, $R_r, R_p, R_y$ denote absolute Euler angles (roll, pitch, yaw) for rotation, and $g \in [0, 1]$ indicates the gripper width. For dual-arm manipulation tasks, the action space is extended to $a_t \in \mathbb{R}^{14}$ by concatenating two 7-DoF vectors corresponding to the left and right arms.

\textbf{VLA Architecture.}
VLA models largely inherit the design of VLMs, consisting of three components: (1) a \textit{vision encoder} that processes multi-view images into visual tokens, (2) a \textit{large language model (LLM)} that models multimodal features, and (3) an \textit{action expert} that generates action predictions, either using the LLM itself or an additional action head. These components are typically initialized from pre-trained VLMs to leverage internet-scale vision–language knowledge.
The training objective maximizes the log-likelihood of action chunks given observations and instructions:
\begin{equation}
    \max_{\theta} \mathbb{E}_{(a_{t:t+H-1}, o_t, \ell) \sim \mathcal{D}} \left[ \log \pi_\theta(a_{t:t+H-1} \mid o_t, \ell) \right].
\end{equation}
The action outputs can be represented as either discrete tokenizations~\cite{2024_9_05_OpenVLA} or continuous formulations, such as diffusion~\cite{2025_3_13_HybridVLA, 2025_5_1_RDT_1B} and flow matching~\cite{2025_3_18_GR00T, 2025_4_22_pi0_5}, which provide more expressive action distributions for complex manipulation tasks.

\section{Proposed Method (Lift3D-VLA)}
\label{sec/method}
Our previous work, Lift3D~\cite{jia2025lift3d}, is limited by its reliance on indirect 3D reconstruction objectives, which hinders effective geometric and physical modeling. Furthermore, it does not jointly model evolving geometry and temporally structured actions, resulting in brittle performance in long-horizon dynamic scenarios.
To address these limitations, as shown in Figure~\ref{fig:1_main_pipeline}, we propose Lift3D-VLA, a unified VLA framework that systematically equips VLA models with explicit 3D point cloud reasoning and facilitates temporally coherent action generation.
We begin by introducing the overall VLA architecture in Section~\ref{sec/vla_architecture}. We then present the 2D model lifting strategy in Section~\ref{sec/3d_lift}, following our previous work Lift3D, enabling VLA models to directly encode 3D point clouds.
In Section~\ref{sec/gc_mae}, we describe our Geometry-Centric Masked Autoencoding (GC-MAE), which enhances the implicit 3D and temporal-aware encoding capabilities of the VLA's 2D encoder.
Building on these representations, we introduce a novel layer-wise temporal action modeling approach in Section~\ref{sec/layer_temporal}, which enables temporally coherent action predictions.
Finally, we present the training objectives in Section~\ref{sec/training_objectives}.

\subsection{Lift3D-VLA Architecture}
\label{sec/vla_architecture}
To enable robots to reason about 3D spatial structures and generate precise manipulation actions, we design a unified VLA architecture that integrates 2D visual observations, 3D point clouds, and language instructions. Our model is built upon a VLM backbone, with parameters initialized from Prismatic VLM~\cite{2024_2_12_Prismatic_VLMs}.
Notably, 2D and 3D observations share a common vision encoder, while employing modality-specific tokenizers and positional embeddings. All raw multisensory inputs are projected into a unified token sequence for the LLM, enabling multimodal reasoning and action prediction.

\textbf{Vision Encoder.}
We utilize a dual-encoder design combining SigLIP~\cite{zhai2023siglip} and DINOv2~\cite{2023_4_14_DINOv2} to capture complementary visual representations, including both global semantic context and fine-grained details. For each input RGB observation $I_t \in \mathbb{R}^{H \times W \times 3}$ (with $H = W = 224$), the shared vision encoder processes the image and extracts dense visual features. This produces feature embeddings $f^{\mathrm{SigLIP}} \in \mathbb{R}^{N_v \times 1024}$ and $f^{\mathrm{DINO}} \in \mathbb{R}^{N_v \times 1152}$, where $N_v = 256$ denotes the number of visual tokens. The resulting features are then concatenated along the channel dimension to form a unified visual representation for downstream processing.

\textbf{3D Point Cloud Tokenizer.}
To augment 2D visual features with explicit 3D geometric information, we introduce a dedicated point cloud tokenizer that converts raw point clouds into structured token representations. The tokenizer consists of three stages: farthest point sampling~\cite{qi2017pointnet} for downsampling, k-nearest neighbors aggregation for capturing local geometric structures, and a learnable linear projection for feature embedding.
Specifically, given a point cloud input $PC \in \mathbb{R}^{1024 \times 3}$, the tokenizer produces a compact representation $f^{\text{pc}} \in \mathbb{R}^{N{\text{pc}} \times d_h}$ with $N_{\text{pc}} = 256$ tokens, where $d_h$ equals the 2D feature channel dimension.
The resulting tokens are subsequently encoded with spatially aligned 3D positional embeddings (described in Section~\ref{sec/3d_lift}) and fed into the shared vision encoder to extract spatial features.

\textbf{LLM Backbone.}
We adopt the 7B LLaMA2 model~\cite{touvron2023llama2} as the LLM backbone for Lift3D-VLA. LLaMA2 follows a decoder-only transformer architecture with 32 layers, where each layer transforms the input token sequence into progressively refined high-dimensional representations.
Visual tokens, geometric tokens, and linguistic tokens are projected into the LLM's word embedding space and concatenated into a unified sequence $f \in \mathbb{R}^{B \times N_t \times d_h}$. This token sequence is then processed jointly by the transformer layers through standard self-attention and feed-forward operations. 
Unlike prior approaches that introduce additional heads after the LLM for action prediction~\cite{2025_4_22_pi0_5, wen2024diffusion} or rely on the final LLM layer for next-token prediction~\cite{2025_3_13_HybridVLA}, we instead leverage sequence representations from intermediate to deep LLM layers to generate temporally coherent action chunks, as detailed in Section~\ref{sec/layer_temporal}.

\subsection{2D Model-lifting Strategy}
\label{sec/3d_lift}
In this section, we propose a lifting mechanism that enables the 2D vision encoder in VLA models to explicitly process point cloud inputs, building upon our previous work, Lift3D~\cite{jia2025lift3d}. Existing approaches typically either project 3D point clouds into multi-view images~\cite{goyal2023rvt, wang2024vihe} or lift 2D features into 3D representations~\cite{gervet2023act3d, shridhar2022cliport}. However, such modality transformations often lead to the loss of geometric structure and spatial consistency, making it challenging to effectively represent 3D data for robotic manipulation. Meanwhile, positional embeddings (PEs) in VLA architectures play a critical role in encoding spatial relationships among tokens. A straightforward solution is to design new 3D positional embeddings, but this may introduce a mismatch with pretrained 2D models and hinder knowledge transfer.

To address this, as illustrated in Figure~\ref{fig:1_main_pipeline} a), we project 3D tokens onto multiple virtual planes and reuse pretrained 2D positional embeddings for 3D encoding. Specifically, raw point clouds are first transformed into high-dimensional features of size $B \times 128 \times 768$ using the 3D point cloud tokenizer. We denote the spatial coordinates of these tokens as $\{C_{3D}^{i}\}_{i=1}^{k}$. Each 3D coordinate is then projected onto $n$ virtual planes, producing corresponding 2D coordinates $\{C_{2D}^{ij}\}_{j=1}^{n}$.
This projection is parameter-free and efficient. We adopt a cube-based projection with six faces to ensure comprehensive coverage of spatial information. In contrast to Lift3D, which relies on randomly assigned front-view virtual planes, we leverage camera extrinsic parameters to align the virtual front view with the observation camera. This alignment enforces a consistent projection viewpoint, improving geometric consistency and reducing distortion.
Each virtual plane shares the pretrained 2D positional embedding grid. Based on the projected coordinates, each 3D token is associated with $n$ positional embeddings, denoted as $\{PE_{2D}(C_{2D}^{ij})\}_{j=1}^{n}$. To obtain a unified positional representation, we aggregate these embeddings by averaging:
\begin{align}
PE_{3D} = \frac{1}{n} \sum_{j=1}^{n} PE_{2D}(C_{2D}^{ij}).
\end{align}
The resulting $PE_{3D}$ is combined with the 3D token features and fed into the 2D vision encoder. By reusing pretrained 2D positional embeddings from multiple views, this design captures diverse spatial relationships while mitigating information loss. With explicit 3D inputs available, we further propose a novel 3D self-supervised training framework to enhance the 3D representations in VLA models.

\subsection{Geometry-Centric Masked Autoencoding}
\label{sec/gc_mae}
In our preliminary work, Lift3D~\cite{jia2025lift3d}, we enhance 2D foundation models with implicit 3D awareness by reconstructing depth maps and RGB from task-relevant patches. While effective for static spatial understanding, this design relies on ``2.5D'' depth representations and does not explicitly enforce coherent 3D geometric structure. Moreover, it treats observations independently, overlooking the temporal dynamics required for continuous manipulation. To address these limitations, as shown in Figure~\ref{fig:1_main_pipeline} b), we propose GC-MAE, a 3D temporal-aware self-supervised learning framework that leverages the 2D Model-lifting Strategy to directly encode point cloud inputs, and jointly models geometric structure and its temporal evolution via a scalable point cloud synthesis pipeline.

\textbf{Scalable 3D Data Synthesis.}
We adopt a scalable 3D data synthesis pipeline to augment existing 2D robotic datasets with 3D point clouds. As most robotic datasets~\cite{o2024open, khazatsky2024droid, hou2025robomind} provide only RGB observations without depth and camera parameters, we leverage VGGT~\cite{2025_3_14_vggt}, a feed-forward transformer pretrained on large-scale visual-geometry data, to generate pseudo 3D annotations. Given an RGB observation $I_t \in \mathbb{R}^{H \times W \times 3}$, VGGT directly predicts the corresponding point cloud $\hat{P}_t \in \mathbb{R}^{N \times 3}$. We apply VGGT to all observations in the pretraining dataset (Table~\ref{tab:data_mix}), producing pseudo point clouds that provide geometric supervision complementary to 2D visual features. We further empirically verify that the quality of these synthesized point clouds is sufficient to support reliable self-supervised training.
Building on the synthesized 3D data, we jointly capture static geometric structure and dynamic evolution by designing a dual-branch decoding framework over shared latent representations.

\textbf{Masked Point Reconstruction.}
To model static geometric structures, we adopt a masked autoencoding paradigm~\cite{2022_3_13_PointMAE} over 3D point tokens. Specifically, a high proportion of input tokens are randomly masked, and the encoder processes only the visible subset $\mathcal{P}_t^{v}$. A lightweight transformer decoder reconstructs the 3D coordinates of masked tokens from encoded visible tokens and learnable mask tokens. This intra-frame geometric completion is supervised using the Chamfer Distance (CD):
\begin{equation}
\label{eq:static_loss}
\mathcal{L}_{\text{static}} =
\sum_{p \in \mathcal{P}_t^{m}}
\min_{\hat{p} \in \hat{\mathcal{P}}_t^{m}}
\lVert p - \hat{p} \rVert_2^2
+
\sum_{\hat{p} \in \hat{\mathcal{P}}_t^{m}}
\min_{p \in \mathcal{P}_t^{m}}
\lVert \hat{p} - p \rVert_2^2,
\end{equation}
where $\mathcal{P}_t^{m}$ denotes the ground-truth point set of masked tokens, and $\hat{\mathcal{P}}_t^{m}$ denotes the reconstructed point set.

\textbf{Future Geometric Prediction.}
To model temporal dynamics, we introduce a future prediction branch that captures inter-frame geometric evolution without explicit motion supervision. Given the visible 3D tokens $\mathcal{P}_{t}^{v}$ at time $t$, the decoder predicts their corresponding geometry in the next frame $\hat{\mathcal{P}}_{t+1}^{v}$. Unlike the static reconstruction objective, this branch enforces temporal causality by learning how geometry evolves over time (e.g., object motion or end-effector trajectories). The temporal loss is defined as:
\begin{equation}
\label{eq:dynamic_loss}
\mathcal{L}_{\text{dynamic}} =
\sum_{p \in \mathcal{P}_{t+1}^{v}}
\min_{\hat{p} \in \hat{\mathcal{P}}_{t+1}^{v}}
\lVert p - \hat{p} \rVert_2^2
+
\sum_{\hat{p} \in \hat{\mathcal{P}}_{t+1}^{v}}
\min_{p \in \mathcal{P}_{t+1}^{v}}
\lVert \hat{p} - p \rVert_2^2.
\end{equation}
To adapt the vision encoder for 3D understanding while preserving large-scale pretrained priors, we inject Low-Rank Adaptation (LoRA)~\cite{2021_6_17_LoRA} into the attention layers. We freeze most backbone parameters and update only the LoRA modules and the 3D tokenizer, ensuring parameter efficiency while mitigating catastrophic forgetting.

\subsection{Layer-wise Temporal Action Modeling}
\label{sec/layer_temporal}

Despite learning strong 3D representations, Lift3D relies on a simple policy head on top of the 2D vision encoder for action prediction, which limits its ability to fully exploit 3D structural reasoning and constrains its generalization capability.
To address this limitation, we leverage the intrinsic sequence modeling capability of Transformer-based LLMs~\cite{chen2021decision, wen2023large} to capture temporal dependencies in physical control. Moreover, recent VLA models have shown that action prediction need not be restricted to the final layer, as intermediate representations often contain rich and actionable information~\cite{yue2024deer, bjorck2025gr00t}.
Building on these insights, as illustrated in Figure~\ref{fig:1_main_pipeline} c), we propose Lift3D-VLA, which introduces a layer-wise temporal action modeling strategy to better utilize the learned spatial features in VLA models while effectively modeling temporal dependencies.
Instead of relying solely on the final layer, we progressively decode action sequences across layers, where each layer predicts a corresponding action step. This layer-wise design naturally captures temporal dependencies, enabling more coherent modeling of action sequences.

Specifically, given an action chunk of horizon $H$ starting from time step $t$, we assign each future step $t+k$ (where $k \in \{0, \dots, H-1\}$) to a corresponding intermediate layer $l_k$ of the LLM backbone.
We then instantiate $H$ action layer $\{\phi_k\}_{k=0}^{H-1}$, where each layer $\phi_k$ predicts the denoised action at time step $t+k$ from the hidden state $\mathbf{h}_{l_k}$:
\begin{equation}
\hat{\boldsymbol{\epsilon}}_k = \phi_k(\mathbf{h}_{l_k}), \quad k \in \{0, \dots, H-1\},
\label{eq:layer}
\end{equation}
where $l_k$ denotes the layer index assigned to step $k$. Ablation studies are conducted to analyze the impact of the number of layers used for action decoding and the interval between decoding layers.
In practice, we empirically find that leveraging the deeper layers of the LLM and uniformly spacing the decoding layers yields the best performance. For example, with a 32-layer LLM backbone and an action chunk of $H=4$, we uniformly select layers ${20, 24, 28, 32}$ from the deeper portion of the network to decode actions, assigning one action step to each selected layer.
More generally, when the action horizon exceeds the number of selected layers, each layer can predict multiple consecutive action steps, enabling flexible and scalable action decoding. 
This hierarchical formulation naturally leverages the temporal conditioning across different layers of the LLM: deeper layers (corresponding to future time steps) inherently attend to features from shallower layers (representing current or near-term states), thereby enabling temporally consistent modeling.
Furthermore, distributing supervision across network depths facilitates the learning of robust features at multiple levels of abstraction, improving closed-loop stability during dynamic interactions.

\subsection{Training Objectives}
\label{sec/training_objectives}

As shown in Figure~\ref{fig:1_main_pipeline} b) and c), our training pipeline consists of two stages: (1) GC-MAE for self-supervised learning, and (2) supervised fine-tuning (SFT) on downstream tasks based on the proposed layer-wise temporal action modeling. After equipping the 2D vision encoder with strong 3D representations in the first stage, the second stage further adapts the VLA model for task-specific action prediction.

\textbf{GC-MAE self-supervised learning.} We optimize the VLA's 2D vision encoder and dual-branch decoder using only geometric supervision, without any action labels. The objective of this stage is a weighted sum of the static reconstruction (Eq.~\ref{eq:static_loss}) and dynamic prediction losses (Eq.~\ref{eq:dynamic_loss}):
\begin{equation}
\mathcal{L}_{\text{MAE}} = \mathcal{L}_{\text{static}} + \lambda \cdot \mathcal{L}_{\text{dynamic}}.
\label{loss:MAE}
\end{equation}
The weighting factor $\lambda$ balances the value of the two losses. By minimizing $\mathcal{L}_{\text{MAE}}$, the encoder learns representations that jointly capture 3D structure and dynamics, thereby equipping the VLA's vision encoder with fundamental 3D understanding for downstream robotic manipulation tasks. Note that the decoders are used only during Stage 1.

\textbf{Supervised Fine-tuning.}
During Stage 2, we freeze the encoder and fine-tune the LLM backbone along with an action projection MLP, which maps the denoised tokens to the action space, using task-specific demonstrations. For action prediction, we adopt a standard DDPM~\cite{ho2020denoising}.
Given the layer-wise temporal action modeling strategy in Eq.~\ref{eq:layer}, we supervise the denoising process simultaneously across all time steps within the action chunk. The action generation loss is defined as the average Mean Squared Error (MSE) over horizon $H$:
\begin{equation}
\mathcal{L}_{\text{action}} =
\mathbb{E}_{k, s, \epsilon} \left[
\left\|
\boldsymbol{\epsilon}_k -
\hat{\boldsymbol{\epsilon}}_\theta(\mathbf{z}_{t+k}^{(s)}, s, \mathbf{h}_{l_k})
\right\|_2^2
\right],
\label{loss:action}
\end{equation}
where $\boldsymbol{\epsilon}_k$ denotes the ground-truth noise added to action $a_{t+k}$, $\mathbf{z}_{t+k}^{(s)}$ is the corresponding noisy action at diffusion timestep $s$, and $\mathbf{h}_{l_k}$ is the hidden representation from layer $l_k$. During inference, we adopt DDIM~\cite{ddim2021} with 4 denoising steps, following prior methods~\cite{liurdt, 2025_3_13_HybridVLA}.

\section{Experiment}
\label{sec/05_experiment}
In Section~\ref{sec/05_experiment/1_implementation_details}, we provide a detailed description of the construction of the self-supervised and pretraining datasets. We then benchmark Lift3D-VLA against recent VLA models in Section~\ref{sec/05_experiment/2_simulation_exp}, evaluating manipulation performance on the simulation environments. The contributions of key components are further isolated and quantified through comprehensive ablation studies in Section~\ref{sec/05_experiment/3_ablation}. Section~\ref{sec/05_experiment/4_real_world_experiment} demonstrates the real-world effectiveness of Lift3D-VLA on manipulation tasks with both single-arm and dual-arm configurations. Finally, in Section~\ref{sec/05_experiment/5_generalization_experiment}, we evaluate the generalization capability of our method under unseen objects, lighting conditions, and scene backgrounds.

\subsection{Pretraining Configuration}
\label{sec/05_experiment/1_implementation_details}

We construct a large-scale pretraining corpus by integrating diverse open-source robotic datasets, including Open-X-Embodiment~\cite{o2024open}, DROID~\cite{khazatsky2024droid}, and RoboMIND~\cite{wu2024robomind}. Detailed statistics on data composition and proportions are provided in Table~\ref{tab:data_mix}. Prior to downstream task fine-tuning, we perform a two-stage pretraining procedure. First, we conduct large-scale robotic data pretraining to endow the VLM with VLA capabilities. Subsequently, we further enhance the vision encoder using our proposed Geometry-Centric MAE (GC-MAE) self-supervised learning framework.

\textbf{For robotic data pretraining}, we curate a diverse corpus of 400K trajectories (28M frames) from the aggregated dataset, enabling the model to acquire a strong foundation of motor primitives and physical commonsense reasoning. During this stage, to ensure supervision accuracy, the model takes only 2D images as input and predicts the end-effector pose as the supervision signal (Eq.~\ref{loss:action}). The implementation details in this stage are consistent with those used for real-world tasks and are described in the following section.

\textbf{For self-supervised learning}, we employ the 3D data synthesis pipeline described in Section~\ref{sec/gc_mae} to generate point cloud annotations from RGB observations, resulting in over 140K trajectories with synthesized 3D annotations. For computational efficiency, point clouds are uniformly subsampled to 1,024 points per frame. This pretraining phase is conducted for 15 epochs on the synthesized 3D dataset, using a batch size of 4096 and a learning rate of $1 \times 10^{-4}$. We adopt the AdamW optimizer with $(\beta_1, \beta_2) = (0.9, 0.999)$, along with a cosine learning rate schedule and linear warmup over the first 10\% of training steps.
We optimize the vision encoder of the VLA model together with the dual-branch decoder (introduced only during training) using the objective defined in Eq.~\ref{loss:MAE}. The decoder comprises separate static and dynamic branches, each consisting of four Transformer layers with eight attention heads per layer.

\begin{table}[t]
    \centering
    \caption{\textbf{Datasets used for pre-training.} The names of selected datasets for large-scale pretraining and their sampling ratios (\%)}
    
    \small 
    \setlength{\tabcolsep}{3pt} 
    
    \begin{tabularx}{\columnwidth}{@{} L r @{\hspace{12pt}} L r @{}}
        \toprule
        \multicolumn{4}{c}{\textbf{\normalsize Training Dataset Mixture}}\\ 
        \midrule
        Fractal          & 9.1\%  & Austin Sirius      & 1.2\% \\
        Kuka        & 27.8\% & CMU Pickup & 0.7\% \\
        Bridge & 4.1\% & UTAustin Mutex & 1.6\% \\
        Taco Play & 2.1\% & Berkeley Fanuc & 0.6\% \\
        Jaco Play    & 0.3\%  & CMU Stretch & 0.1\% \\
        Berkeley Cable & 0.2\%  & BC-Z                 & 5.4\% \\
        Roboturk & 1.7\% & FMB Dataset & 5.0\% \\
        Viola           & 0.7\%  & DobbE       & 1.0\% \\
        Berkeley UR5  & 0.9\%  & DROID        & 7.2\% \\
        Toto            & 1.5\%  & Stanford Kuka & 0.1\% \\
        Language Table & 3.1\% & Robocook & 0.1\% \\
        Stanford Hydra & 3.2\% & Maniskill & 6.3\% \\
        Austin Buds    & 0.2\%  & Berkeley RPT & 0.1\% \\
        NYU Franka       & 0.6\%  & QUT Dexterous   & 0.1\% \\
        Furniture Bench & 1.8\% & RoboSet & 1.8\% \\
        UCSD Kitchen   & $<$0.1\% & BridgeData V2 & 4.7\% \\
        Austin Sailor & 1.6\% & RoboMind       & 5.2\% \\
        \bottomrule
    \end{tabularx}
    \label{tab:data_mix}
\end{table}

\subsection{Simulation Experiment}
\label{sec/05_experiment/2_simulation_exp}

\begin{table*}[h]
\caption{\textbf{Results on the MetaWorld.}
We evaluate models in both single-task and multi-task settings over 25 rollouts. Success is determined by the built-in MetaWorld evaluation module. Results report average manipulation success rates (S.R.). 
}
\centering
\small
\resizebox{\textwidth}{!}{
\begin{tabular}{l|ccccccccccccc|c}
    \toprule
    \multirow{2}{*}{Models} & \multirow{2}{*}{Assembly} & Bin & Box & Button & Dial & Drawer & \multirow{2}{*}{Hammer} & Hand & Peg & \multirow{2}{*}{Push} & Reach & Shelf & \multirow{2}{*}{Sweep} & \multirow{2}{*}{\makecell{Mean \\ S.R.}} \\
    & & picking & close & press & turn & open & & unplug & wall & & place & into & & \\
    \midrule
    \multicolumn{15}{c}{\textit{Single-Task Setting}} \\
    \midrule
    CLIP~\cite{2021_2_26_CLIP} & 64 & 92 & 60 & 100 & 82 & 100 & 88 & 36 & 78 & 64 & 56 & 26 & 40 & 68.2 \\
    R3M~\cite{2022_11_18_R3M} & 100 & 60 & 92 & 92 & 100 & 100 & 60 & 66 & 96 & 60 & 60 & 36 & 60 & 75.5 \\
    VC-1~\cite{2023_3_31_VC_1} & 60 & 80 & 66 & 96 & 76 & 100 & 88 & 44 & 50 & 60 & 60 & 36 & 60 & 67.4 \\
    PointNet~\cite{qi2017pointnet} & 100 & 44 & 46 & 100 & 94 & 100 & 38 & 32 & 68 & 36 & 52 & 14 & 24 & 57.5 \\
    PointNet++~\cite{2017_6_PointNet_plus_plus} & 96 & 72 & 86 & 98 & 78 & 84 & 70 & 26 & 78 & \textbf{98} & 48 & 12 & 24 & 66.9 \\
    PointNeXt~\cite{2022_6_09_PointNeXt} & 98 & 82 & 78 & 100 & 92 & 100 & 50 & 20 & 78 & 28 & 48 & 12 & 42 & 63.7 \\
    SPA~\cite{zhu2024spa} & 96 & 92 & 76 & 100 & 84 & 96 & 100 & 36 & 68 & 55 & 56 & 16 & 64 & 72.2 \\
    DP3~\cite{ze20243d} & 100 & 24 & 48 & 100 & 92 & 100 & 100 & 14 & 98 & 54 & 40 & 18 & 22 & 62.3 \\
    \midrule
    Lift3D(Clip) & 100 & 92 & 92 & 100 & 100 & 100 & 94 & 64 & 98 & 44 & 74 & 42 & 72 & 82.5 \\
    Lift3D(Dinov2) & 100 & \textbf{100} & 92 & 100 & 100 & 100 & 100 & \textbf{76} & 96 & 40 & 80 & 28 & 80 & 84.0 \\
    \rowcolor[HTML]{FFF0F5}
    \textbf{Lift3D-VLA$^\dagger$(Clip)} & \textbf{100} & 92 & \textbf{100} & 100 & 100 & 100 & 92 & 60 & \textbf{100} & 56 & 88 & \textbf{72} & 92 & \textbf{88.6} \\
    \rowcolor[HTML]{FFF0F5}
    \textbf{Lift3D-VLA$^\dagger$(Dinov2)} & 96 & 92 & 100 & \textbf{100} & \textbf{100} & \textbf{100} & \textbf{100} & 56 & 96 & 40 & \textbf{92} & 68 & \textbf{92} & 87.1 \\
    \midrule
    \multicolumn{15}{c}{\textit{Multi-Tasks Setting}} \\
    \midrule
    OpenVLA~\cite{2024_9_05_OpenVLA} & 90 & 70 & \textbf{70} & 100 & 90 & 100 & 50 & 70 & 80 & 50 & 30 & 70 & \textbf{90} & 73.9 \\
    $\pi_{0.5}$~\cite{2025_4_22_pi0_5} & 100 & 85 & 50 & 100 & 45 & 80 & 75 & 45 & 65 & 55 & 45 & \textbf{75} & 60 & 67.7 \\
    SpatialVLA~\cite{qu2025spatialvla} & 75 & 65 & 50 & 90 & 65 & 65 & 85 & 45 & 55 & 35 & 65 & 40 & 70 & 61.9 \\
    3DS-VLA~\cite{2025_8_08_3DS_VLA} & 75 & 100 & 45 & 90 & 95 & 100 & 90 & 70 & 75 & 50 & 60 & 60 & \textbf{90} & 76.9 \\
    \midrule
    \rowcolor[HTML]{FFF0F5}
    \textbf{Lift3D-VLA} & \textbf{100} & \textbf{100} & 64 & \textbf{100} & \textbf{100} & \textbf{100} & \textbf{96} & \textbf{88} & \textbf{100} & \textbf{84} & \textbf{72} & 52 & 84 & \textbf{87.7} \\
    \bottomrule
\end{tabular}}
\label{tab:2_1_single_and_multi_tasks_metaworld}
\end{table*}

\begin{table*}[t]
\caption{\textbf{Results on the RLBench.}
All models are trained in the multi-task setting and evaluated over 25 rollouts.
}
\centering
\resizebox{\textwidth}{!}{ 
\begin{tabular}{l|ccccccccc|c}
    \toprule
    \multirow{2}{*}{Models} & Close & Close & Toilet & Sweep to & Close  & Umbrella  & Frame & Place wine & Water  & Mean \\
                            & box   & laptop lid & seat down & dustpan & fridge &  out & off hanger & at rack & plants & S.R. \\
    \midrule
    OpenVLA~\cite{2024_9_05_OpenVLA}  & 60 & 35 & 75 & 55 & 85 & 30 & 15 & 20 & 5 & 42.2 \\
    $\pi_{0.5}$~\cite{2025_4_22_pi0_5} & 90 & 95 & 85 & 75 & \textbf{100} & 10 & \textbf{80} & 75 & 35 & 71.7 \\
    SpatialVLA~\cite{qu2025spatialvla} & 80 & 70 & 85 & 20 & 80 & 25 & 40 & 15 & 30 & 49.4 \\
    3DS-VLA~\cite{2025_8_08_3DS_VLA} & 85 & \textbf{95} & \textbf{95} & 15 & 90 & \textbf{80} & 50 & 85 & 35 & 70.0 \\
    \midrule
    \rowcolor[HTML]{FFF0F5}
    \textbf{Lift3D-VLA} & \textbf{95} & 80 & \textbf{95} & \textbf{95} & 90 & 65 & \textbf{80} & \textbf{95} & \textbf{50} & \textbf{82.8} \\
    \bottomrule
\end{tabular}}
\label{tab:2_2_multi_tasks_rlbench}
\end{table*}

\subsubsection{Data Collection}
We evaluate our method on two commonly used simulation environments. \textbf{MetaWorld}~\cite{2019_10_24_Meta_World} provides a large-scale suite of simulated manipulation tasks designed to assess the robustness of robotic policies. In our experiments, we follow~\cite{jia2025lift3d} and select 13 tasks that cover a diverse set of manipulation skills, including precise positioning, articulated object interaction, and fine-grained manipulation: 1) \textit{Assembly}, 2) \textit{Bin picking}, 3) \textit{Box close}, 4) \textit{Button press}, 5) \textit{Dial turn}, 6) \textit{Drawer open}, 7) \textit{Hammer}, 8) \textit{Hand insert}, 9) \textit{Peg unplug side}, 10) \textit{Push wall}, 11) \textit{Reach}, 12) \textit{Shelf place}, and 13) \textit{Sweep into}. All experiments are conducted using a Sawyer robotic arm equipped with a parallel-jaw gripper. For each task, we collect 100 trajectories using scripted policies provided by the benchmark. All RGB observations are captured from a single third-view corner camera at a resolution of $224 \times 224$.
\textbf{RLBench}~\cite{2019_9_26_RLBench} is a manipulation benchmark built on the CoppeliaSim robotics simulator. In this environment, we follow previous VLA papers~\cite{2025_3_13_HybridVLA, liu2026last} and select 9 tasks that span a range of manipulation skills: 1) \textit{Close box}, 2) \textit{Close laptop}, 3) \textit{Toilet seat down}, 4) \textit{Sweep to dustpan}, 5) \textit{Close fridge}, 6) \textit{Take umbrella out}, 7) \textit{Frame off hanger}, 8) \textit{Wine at rack}, and 9) \textit{Water plants}. All tasks are performed using a Franka Panda robotic arm with a single front-view camera observation. Demonstration trajectories are generated using the Open Motion Planning Library (OMPL)~\cite{sucan2012open}, with 100 trajectories per task, and keyframes are extracted following prior work~\cite{goyal2023rvt, 2022_11_11_PerAct_ERCEIVER_ACTOR}.
3D point cloud annotations are uniformly subsampled to 1,024 points per frame.

\subsubsection{Single-Task on MetaWorld}
\label{sec/05_experiment/2_simulation_exp/single_task_setting}
Since our previous work, Lift3D~\cite{jia2025lift3d}, does not incorporate language understanding, it cannot be applied to multi-task manipulation. To enable a fair comparison and better evaluate the effectiveness of our proposed method, we validate the GC-MAE self-supervised learning strategy in a single-task setting. Specifically, consistent with Lift3D, we attach an MLP head to the pretrained vision encoder to predict actions, instead of using the full LLM in Lift3D-VLA for action prediction.

\textbf{Baselines.} For 2D visual representation learning, we include CLIP (ViT-Base)~\cite{2021_2_26_CLIP}, R3M~\cite{2022_11_18_R3M}, and VC~\cite{2023_3_31_VC_1}, all of which are pretrained on large-scale vision datasets and adapted for robotic control. For 3D representation learning, we consider PointNet~\cite{qi2017pointnet}, PointNet++~\cite{2017_6_PointNet_plus_plus}, PointNeXt~\cite{2022_6_09_PointNeXt}, and SPA~\cite{zhu2024spa}, which include both generic point cloud encoders and 3D robotic pretraining methods. For 3D robotic policies, we compare with DP3~\cite{ze20243d} and our previous work, Lift3D.

\textbf{Implementation Details.}
We denote our method as Lift3D-VLA$^\dagger$, which shares identical training configurations with the prior Lift3D model. Both models are trained using the same setups, including a three-layer MLP policy head, the Adam optimizer, and a constant learning rate of $1 \times 10^{-3}$. During downstream fine-tuning, the backbone parameters are frozen, and LoRA~\cite{2021_6_17_LoRA} with rank $r=2$ is applied to the attention layers. All models are trained for 100 epochs on 8 NVIDIA A800 GPUs. We conduct 25 rollout evaluations every 5 epochs and report the highest average success rate.

\textbf{Quantitative Results.} 
As shown in the single-task results in Table~\ref{tab:2_1_single_and_multi_tasks_metaworld}, incorporating our GC-MAE pretraining consistently improves performance across different visual backbones. Compared with prior baselines, our model significantly outperforms VC-1 (67.4\%), SPA (72.2\%), and DP3 (62.3\%), demonstrating the effectiveness of combining large-scale 2D pretrained knowledge with robust 3D representations. Notably, Lift3D-VLA$^\dagger$ consistently surpasses Lift3D under both CLIP and DINOv2 initialization, achieving 88.6\% vs. 82.5\% and 87.1\% vs. 84.0\%, respectively.
The gains are particularly evident in tasks requiring sustained spatial precision: shelf-place (72\% vs.\ 42\%), sweep-into (92\% vs.\ 72\%), and reach (92\% vs.\ 80\%).
These results highlight not only the effectiveness of directly reconstructing 3D point clouds, but also the importance of jointly modeling their future geometric evolution for manipulation tasks.

\subsubsection{Multi-Task on MetaWorld and RLBench}
\label{sec/05_experiment/2_simulation_exp/multi_tasks_setting}
In this section, we conduct multi-task experiments. 

\textbf{Baselines.}
Since our previous work, Lift3D, does not support understanding task instructions, it cannot be applied to multi-task settings. Therefore, we compare Lift3D-VLA with recent state-of-the-art (SOTA) VLA models.
For 2D VLA baselines, OpenVLA~\cite{2024_9_05_OpenVLA} leverages a large pretrained vision-language model to perform end-to-end autoregressive action prediction. $\pi_{0.5}$~\cite{2025_4_22_pi0_5} is trained on a large-scale, self-collected robotic dataset, enabling strong generalization.
For 3D VLA baselines, SpatialVLA~\cite{qu2025spatialvla} augments VLA models with explicit spatial reasoning modules to capture geometric structure, while 3DS-VLA~\cite{2025_8_08_3DS_VLA} incorporates 3D structural constraints to further enhance spatial understanding.

\textbf{Implementation Details.}
All baselines are initialized from their officially released pretrained checkpoints and trained using their original full fine-tuning configurations. For Lift3D-VLA, we also adopt full fine-tuning and jointly train across all tasks for 300 epochs with a learning rate of $1 \times 10^{-4}$. All remaining settings are consistent with the single-task setup.
Evaluation follows the protocol of prior work~\cite{chen2025fast}. We perform 20 rollouts per task using the final checkpoint, repeat the evaluation with three different random seeds, and report the average success rate.

\textbf{Quantitative Results.} 
In multi-task experiments, Lift3D-VLA achieves 87.7\% average success across 13 MetaWorld tasks, substantially outperforming OpenVLA, $\pi_{0.5}$, SpatialVLA, and 3DS-VLA by 13.8\%, 20.0\%, 25.8\%, and 10.8\%, respectively. Notably, our method surpasses our previous work, even though Lift3D is trained in a single-task setting, demonstrating both stronger spatial reasoning and improved model capacity. On RLBench (Table~\ref{tab:2_2_multi_tasks_rlbench}), Lift3D-VLA achieves 82.8\% average success across 9 tasks, consistently outperforming prior VLA methods. In particular, it attains near-perfect success rates on tasks requiring precise spatial reasoning, such as \textit{close box} (95\%), \textit{toilet seat down} (95\%), and \textit{place wine at rack} (95\%). These results highlight the effectiveness of combining explicit 3D spatial grounding with temporally coherent action generation.

\begin{figure}[t]
    \centering
    \includegraphics[width=0.49\textwidth]{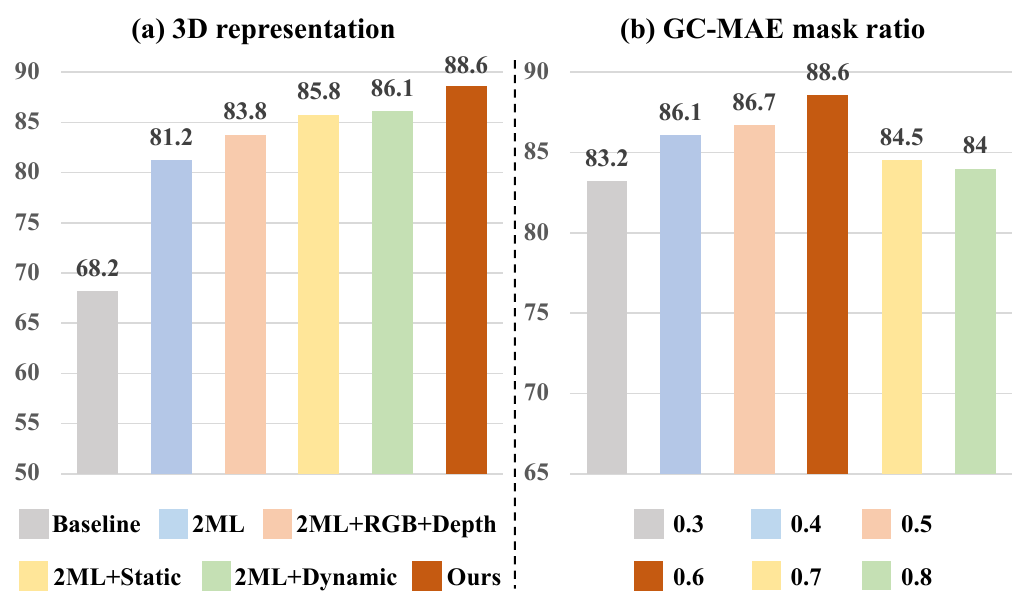}
    \caption{\textbf{Ablation Studies.} 
    \textbf{(a)} Impact of the 2D model-lifting strategy and GC-MAE on 3D representation.
\textbf{(b)} Effect of the mask ratio in GC-MAE.
All experiments are conducted in the MetaWorld single-task setting.}
    \label{fig:ablation_study_combined}
\end{figure}

\begin{table}[t]
\centering
\caption{\textbf{Ablation of layer-wise temporal action modeling.} `Layers' indicates the number of Transformer layers used for parallel action prediction, and `Stride' denotes the interval between them.}
\label{tab:ablation_stride_major}
\small
\resizebox{0.99\linewidth}{!}{%
\begin{tabular}{l c c c}
\toprule
\multirow{2}{*}[-\dimexpr\ht\strutbox/2]{\textbf{Metric}} & \textbf{1 Layer} & \textbf{2 Layers} & \textbf{4 Layers} \\
\cmidrule(lr){2-2} \cmidrule(lr){3-3} \cmidrule(lr){4-4}
& Stride 1 & Stride 1 / 2 / 4 & Stride 1 / 2 / 4 \\
\midrule
\textbf{Mean S.R.} & 82.5 & 84.3 / 85.1 / 85.8 & 84.0 / 86.3 / \textbf{87.7} \\
\bottomrule
\end{tabular}
}
\end{table}
\subsection{Ablation Study}
\label{sec/05_experiment/3_ablation}

We conduct comprehensive ablation studies to validate the effectiveness of each component in Lift3D-VLA. 

\subsubsection{3D Representation Analysis}
\label{sec/05_experiment/3_ablation/component}

As shown in Figure~\ref{fig:ablation_study_combined}(a), we systematically ablate the 2D model-lifting strategy and GC-MAE to evaluate their impact on the 3D representation of Lift3D-VLA in the single-task MetaWorld setting (using MLP action head). Without any 3D modeling, the RGB \textit{baseline} achieves only 68.2\%, highlighting the necessity of geometric reasoning. Applying our enhanced 2D model-lifting strategy (\textit{2ML}), which leverages camera parameters to define a virtual front view, improves performance to 81.2\%, indicating that better geometric alignment between 3D points and 2D positional embeddings provides more effective spatial grounding. Building on this explicit 3D representation, incorporating the implicit \textit{RGB+Depth} reconstruction objective from our previous Lift3D further improves performance to 83.8\%. We then replace the reconstruction objective with point cloud self-supervised learning. Using only the static branch for masked point reconstruction (\textit{2ML+Static}) achieves 85.8\%, while using only the dynamic branch for future geometric prediction (\textit{2ML+Dynamic}) yields 86.1\%. The full GC-MAE model (Ours) achieves 88.6\%, demonstrating the complementary roles of the two branches in enabling robust 3D understanding of physical dynamics.

\begin{figure}[t]
    \centering
    \includegraphics[width=0.49\textwidth]{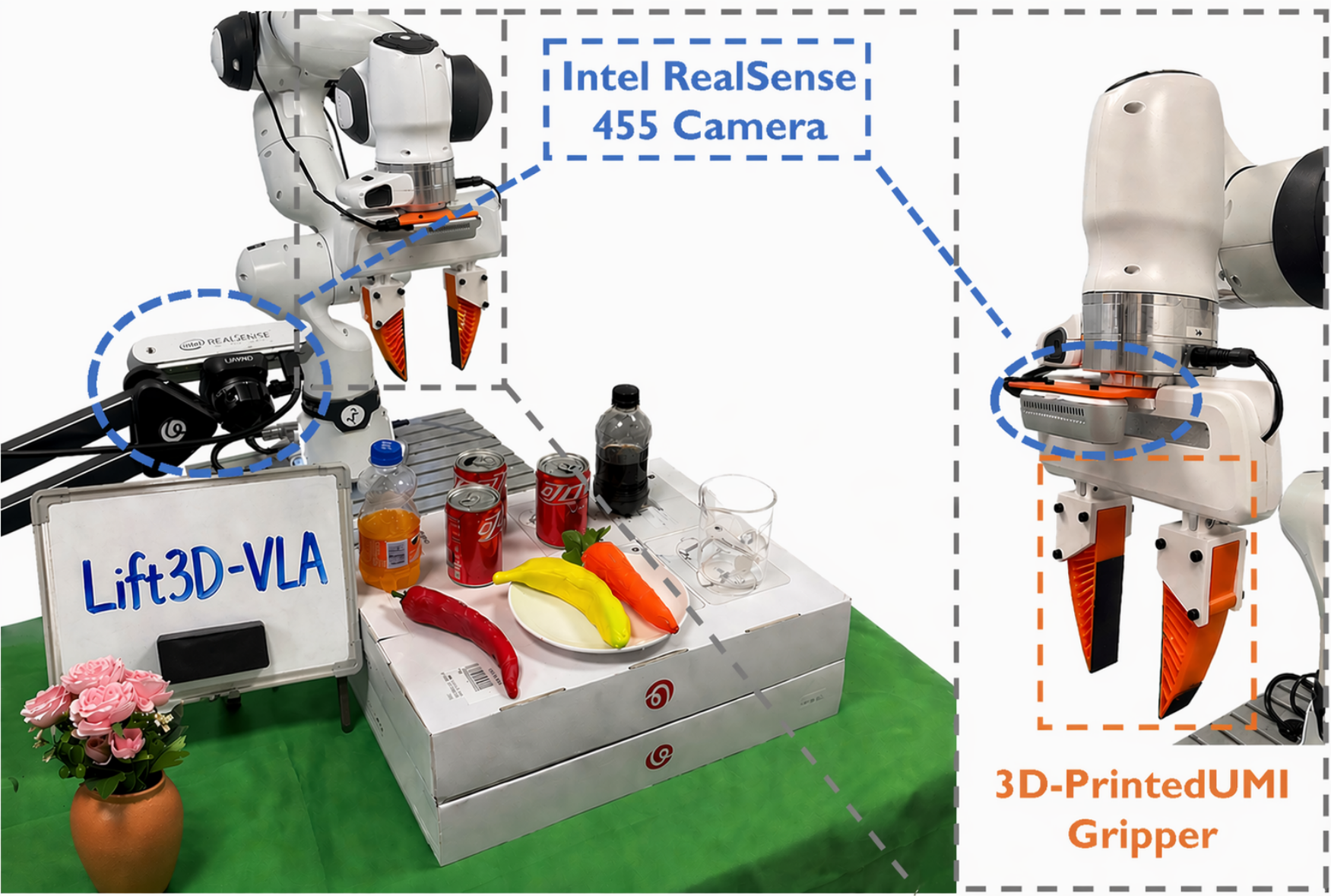}
    \caption{        \textbf{Real-world experimental robot platform.}
        We employ a Franka Research~3 arm equipped with an Intel RealSense D455 RGB-D camera.}
    \label{fig:exp_setup_single_arm}
\end{figure}

\subsubsection{GC-MAE Hyperparameter Analysis}
\label{sec/05_experiment/3_ablation/hyperparameter}

We examine three critical hyperparameters in GC-MAE pretraining: the mask ratio, decoder depth, and the scale of the pretraining data. \textbf{Mask ratio.} As shown in Figure~\ref{fig:ablation_study_combined} (b), a mask ratio of 0.6 yields the best performance. This is because our framework jointly performs masked point cloud reconstruction and future prediction on visible regions, requiring a moderate mask ratio to balance the two objectives.
\textbf{Decoder Depth.} 
We vary the depth of the MAE decoder by setting it to 1, 2, 4, and 16 layers, achieving 84.3\%, 86.1\%, 86.6\%, and 87.2\%, respectively. Both 4-layer and 16-layer decoders yield strong performance, indicating that a relatively lightweight decoder is sufficient for GC-MAE self-supervised learning.
\textbf{Pretraining data.} 
We vary the size of the pretraining corpus to analyze its impact on downstream manipulation performance. Using 20K and 140K samples for GC-MAE pretraining yields 85.6\% and 88.6\% success rates, respectively, demonstrating the scalability of our method with increased data.

\subsubsection{Layer-wise Temporal Action Modeling Analysis}

We further investigate the design of the layer-wise temporal action modeling strategy on the MetaWorld multi-task setting, providing insights into the formulation of action chunk generation. As shown in Table~\ref{tab:ablation_stride_major}, \textit{Layers} denotes the number of layers used for action prediction, while \textit{Stride} indicates the interval between selected layers. For example, \textit{Layers}=1 and \textit{Stride}=1 correspond to predicting all action chunks from the final layer, whereas \textit{Layers}=4 and \textit{Stride}=4 (for a 32-layer LLM) use layers 20, 24, 28, and 32, with each layer predicting a subset of consecutive action steps.
For clarity, all experiments use a fixed action chunk size of four.
Using a single layer yields a baseline success rate of 82.5\%. Incorporating additional intermediate LLM layers (e.g., two or four layers) for action chunk prediction consistently improves performance, indicating that leveraging intermediate representations enhances robustness. Moreover, given the same number of layers, larger stride intervals between LLM output layers lead to further performance gains, suggesting the benefit of capturing more temporally diverse features. These results validate our layer-wise action modeling design, where parallel predictions from intermediate LLM layers enable richer temporal modeling of actions and improve downstream manipulation performance.

\subsection{Real-World Experiment}
\label{sec/05_experiment/4_real_world_experiment}

To evaluate the practical applicability of Lift3D-VLA, we instantiate our framework on real-world robot platforms using both single-arm  and dual-arm Franka Research~3 setups.

\begin{table*}[t]
\caption{\textbf{Comparison across real-world manipulation tasks.} 
We report success rates (S.R.) for standard single-arm and dual-arm tasks (Franka). Mean S.R. denotes the average success rate.
}
\label{tab:real_world_results}
\centering
\small
\resizebox{\textwidth}{!}{ 
\begin{tabular}{l|cccccccc|c|ccc}
    \toprule
    \multirow{2}{*}{Models} & Wipe & Place dish & Place egg & Pick and  & Pour water  & Stack  & Scoop & Open pot& Mean & \multicolumn{3}{c}{Place egg on bread}\\
                            & whiteboard   & on rack & on bread & place banana & into cup &  cola cans & popcorn & pick corn & S.R. & Step1 & Step2 & Step3\\
    \midrule
    SpatialVLA~\cite{qu2025spatialvla}  & 60 & 33 & 20 & 40 & 87 & 33 & 27 & 40 & 43 & 20 & 7 & 0\\
    $\pi_{0.5}$~\cite{2025_4_22_pi0_5} & 60 & 60 & 47 & \textbf{87} & 87 & \textbf{66} & \textbf{53} & 60 & 65 & 47 & 20 & 7\\
    CoT-VLA~\cite{zhao2025cot} & 53 & \textbf{66} & 33 & 53 & 47 & 33 & 33 & 53 & 46 & 33 & 13 & 7 \\
    \midrule
    \rowcolor[HTML]{FFF0F5}
    \textbf{Lift3D-VLA} & \textbf{66} & \textbf{66} & \textbf{66} & \textbf{87} & \textbf{93} & \textbf{66} & 47 & \textbf{73} & \textbf{71} & \textbf{66} & \textbf{33} & \textbf{20}\\
    \bottomrule
\end{tabular}}
\end{table*}

\begin{figure*}[t]
    \centering
    \includegraphics[width=0.99\textwidth]{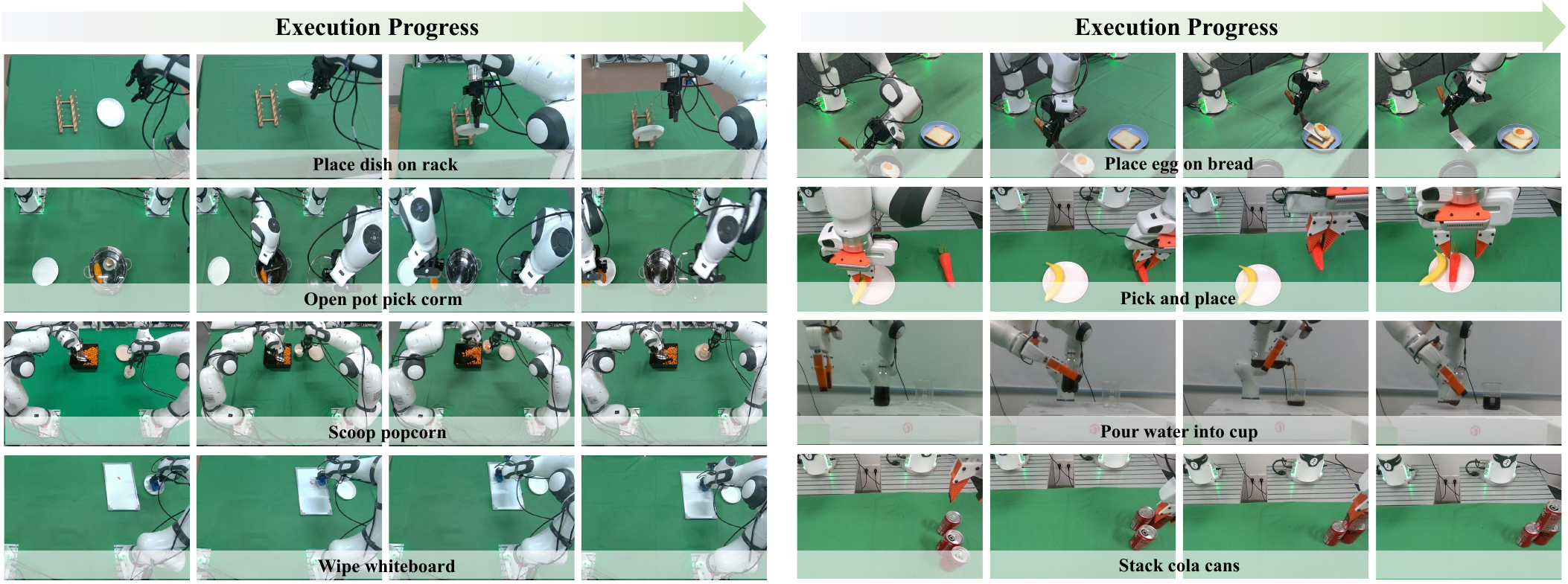}
    \caption{
       \textbf{Visualization.} We illustrate the execution progress of all real-world task execution.
    }
    \label{fig:real_world_execution_progress}
\end{figure*}

\subsubsection{Data Collection}

As shown in Figure~\ref{fig:exp_setup_single_arm}, we illustrate all task assets and real-world experimental setups. In the single-arm configuration, we curate a dataset of 200 demonstrations per task across six manipulation skills: (1) \textit{whiteboard cleaning with an eraser}, (2) \textit{precision placement of a dish onto a rack}, (3) \textit{placing an egg on bread using a spatula}, (4) \textit{picking up a banana and placing it on a white plate}, (5) \textit{pouring water into a cup}, and (6) \textit{stacking three cola bottles}. 
Meanwhile, we evaluate a long-horizon \textit{placing egg} task, where the robot repeats the task three times under continuously changing object positions.
The system is equipped with two Intel RealSense cameras, including one third-person view and one wrist view.
To evaluate dual-arm coordination, we design two collaborative tasks: (1) \textit{scooping popcorn into a bowl}, where one arm stabilizes the bowl while the other performs the scooping motion, and (2) \textit{opening a pot lid and picking corn}, which requires sequential coordination between both arms. In this setting, we use three Intel RealSense cameras, including one third-person view and two wrist views. Point clouds are obtained by projecting third-person depth maps using camera parameters. All trajectories are recorded at 30 FPS via human teleoperation using the 3D space mouse.

\subsubsection{Training and Evaluation Protocol}

We train Lift3D-VLA following the protocol described in Section~\ref{sec/05_experiment/2_simulation_exp/multi_tasks_setting}, with the primary difference being the use of multi-view visual inputs: two camera perspectives for single-arm tasks and three perspectives for dual-arm tasks. We compare our method against three representative baselines: (i) $\pi_{0.5}$~\cite{2025_4_22_pi0_5}, a state-of-the-art 2D VLA model; (ii) SpatialVLA~\cite{qu2025spatialvla}, a 3D-aware VLA framework; and (iii) CoT-VLA~\cite{zhao2025cot}, which incorporates explicit chain-of-thought reasoning for action generation. 
For fair comparison, all methods are initialized from official pretrained checkpoints, trained with full fine-tuning, and evaluated over 15 rollouts per task under varying object positions.

\subsubsection{Quantitative and Qualitative Results}
Table~\ref{tab:real_world_results} summarizes the real-world manipulation performance of Lift3D-VLA and competitive baselines. Our framework achieves an average success rate of 71\% across all tasks, outperforming $\pi_{0}$ (65\%), SpatialVLA (43\%), and CoT-VLA (46\%).
In the \textit{placing egg on bread} task, which involves tool use and complex contact dynamics, Lift3D-VLA achieves a success rate of 66\%, largely attributed to our recover–forecast 3D representation, which enables the robot to reason about geometric interactions between objects.
We further evaluate Lift3D-VLA on a long-horizon manipulation task requiring one, two, and three consecutive successful executions within a single rollout. 
Lift3D-VLA maintains markedly higher success rates across all stages (66\% $\to$ 33\% $\to$ 20\%) compared to $\pi_{0.5}$ (47\% $\to$ 20\% $\to$ 7\%), with the performance gap widening as the horizon increases. This demonstrates that our proposed layer-wise temporal action modeling enables more coherent action generation in long-horizon tasks.
For dual-arm tasks, Lift3D-VLA achieves the highest average success rate of 60\%. For example, in the \textit{opening a pot lid and picking corn} task, our method achieves a 73\% success rate compared to 60\% for $\pi_{0.5}$, demonstrating that improved 3D structural understanding enables more effective temporally coordinated dual-arm actions.
As shown in Figure~\ref{fig:real_world_execution_progress}, we visualize the execution progress across several tasks in both single-arm and dual-arm configurations. The results demonstrate that our method can execute tasks accurately and smoothly across a wide range of scenarios, including basic manipulation, contact-rich interactions, tool use, and structured stacking. Additional execution videos are provided on our project website.

\begin{figure}[t]
    \centering
    \includegraphics[width=0.95\columnwidth]{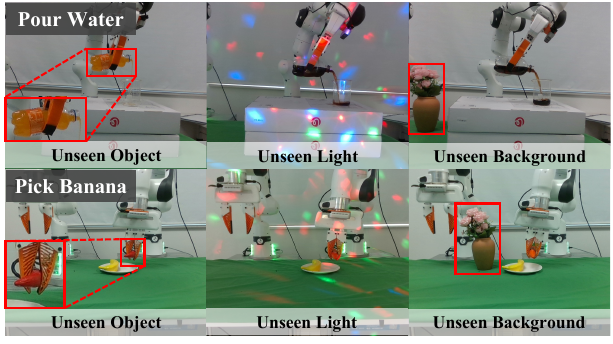}
    \caption{
       \textbf{Generalization.} We evaluate the model under three OOD conditions: \textit{Unseen Object}, \textit{Unseen Light}, and \textit{Unseen Background}. Red boxes highlight the changed conditions.
    }
    \label{fig:generalization}
\end{figure}

\begin{table}[t]
\centering
\caption{\textbf{Generalization experiments.} We evaluate the robustness of \textbf{Lift3D-VLA} against $\pi_{0.5}$ under three unseen scenarios. The percentages in brackets denote the performance degradation relative to the \textit{Original} setup.}
\label{tab:generalization_results}
\vspace{2pt}
\small
\resizebox{\linewidth}{!}{
\begin{tabular}{l cc cc}
\toprule
\multirow{2}{*}{\textbf{Scenario}} & \multicolumn{2}{c}{\textbf{Pick Banana}} & \multicolumn{2}{c}{\textbf{Pour Water}} \\
\cmidrule(lr){2-3} \cmidrule(lr){4-5}
& $\pi_{0.5}$ & Lift3D-VLA & $\pi_{0.5}$ & Lift3D-VLA \\
\midrule
Original & 87 & 87 & 87 & 93 \\
\midrule
Unseen Object       & 80 \scriptsize{(-8\%)}  & 87 \scriptsize{(-0\%)}  & 80 \scriptsize{(-8\%)}  & 87 \scriptsize{(-6\%)} \\
Unseen Lighting     & 73 \scriptsize{(-16\%)} & 80 \scriptsize{(-8\%)}  & 67 \scriptsize{(-23\%)} & 80 \scriptsize{(-14\%)} \\
Unseen Background   & 47 \scriptsize{(-46\%)} & 80 \scriptsize{(-8\%)}  & 60 \scriptsize{(-31\%)} & 87 \scriptsize{(-6\%)} \\
\midrule
\rowcolor[HTML]{FFF0F5}
\textbf{Average Drop} & 64(-26\%) & 82(-6\%) & 69(-21\%) & 85(-9\%) \\
\bottomrule
\end{tabular}}
\end{table}

\subsection{Generalization Experiment}
\label{sec/05_experiment/5_generalization_experiment}

To evaluate the robustness of Lift3D-VLA, we conduct experiments under three out-of-domain (OOD) settings: (1) \textit{Unseen Object}, where target objects (e.g., the bottle or banana) are replaced with different instances; (2) \textit{Unseen Lighting}, where multi-colored lighting perturbations are introduced; and (3) \textit{Unseen Background}, where novel distractor objects (e.g., a flower vase) are added to the scene. Example scenarios are shown in Figure~\ref{fig:generalization}. We compare against the strongest baseline in real-world experiments, $\pi_{0.5}$.
As shown in Table~\ref{tab:generalization_results}, $\pi_{0.5}$ exhibits performance degradation across all OOD settings, particularly under \textit{Unseen Background}, where performance drops by 46\% on \textit{Pick Banana} (87$\rightarrow$47). This sharp decline suggests a strong reliance on 2D appearance and sensitivity to background clutter. In contrast, Lift3D-VLA maintains stable performance across all scenarios, with performance drops consistently bounded within 6\%–8\%.  These results demonstrate that incorporating robust 3D representations significantly improves the model’s understanding of object relationships and enhances generalization.
Under \textit{Unseen Object}, Lift3D-VLA shows minimal degradation (0\% on \textit{Pick Banana}), indicating strong invariance to object appearance. Under \textit{Unseen Lighting}, the performance gap further widens, suggesting that GC-MAE pretraining encourages the model to focus on geometric structure rather than being affected by pixel-level perturbations.
Overall, the consistently smaller performance drops across tasks validate that combining explicit 3D reasoning with temporal action modeling leads to stronger generalization in dynamic manipulation scenarios.

\begin{figure}[t]
    \centering
    \includegraphics[width=1.0\columnwidth]{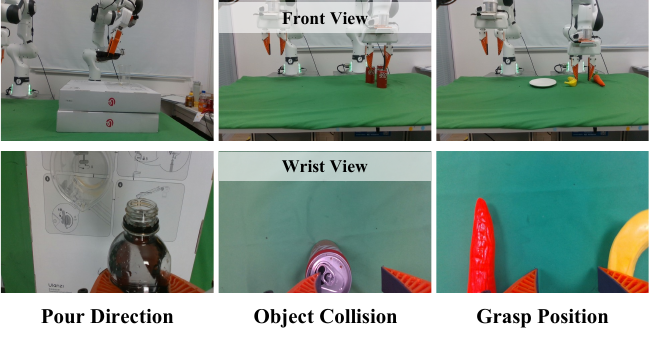}
    \caption{\textbf{Failure case visualization in real-world tasks.} The top row shows the Front View, and the bottom row shows the corresponding Wrist View.}
    \label{fig:failure_cases}
\end{figure}

\subsection{Failure Case Analysis}
\label{sec/05_experiment/4_real_world_experiment/failure_case_analysis}

Despite the strong performance of Lift3D-VLA, we observe several failure cases during real-world execution, as illustrated in Figure~\ref{fig:failure_cases}.

1) \textit{Pour Direction Misalignment}: In the water-pouring task, the model occasionally fails to precisely align the bottle opening with the beaker. This is primarily due to the narrow tolerance of the target container and the limitations of depth sensing on transparent objects, which result in incomplete point cloud observations.

2) \textit{Object Collision}: During the descent phase in the cola-stacking task, the gripper may prematurely collide with the side of the can. This suggests that, despite strong geometric priors from our method, single-view point clouds remain insufficiently robust under certain viewpoints.

3) \textit{Inaccurate Grasp Position}: In the fruit-picking task, the gripper sometimes moves toward the midpoint between two adjacent fruits rather than accurately localizing a single target. This indicates insufficient instance-level discrimination when multiple similar objects are closely positioned.

Overall, these failure cases highlight that, while our method provides robust 3D representations and temporally coherent action modeling, depth perception remains limited in scenarios involving transparent or highly reflective objects. Furthermore, although single-view point clouds are efficient and practical, they lack complete 3D information. In future work, we plan to explore depth completion techniques and multi-view fusion to further improve perception robustness.

\section{Conclusion and Limitation}
\label{sec/06_conclusion}

In this paper, we introduced Lift3D-VLA, a unified framework that integrates explicit 3D point-cloud reasoning with temporally structured action generation for robotic manipulation. Building upon the 2D model lifting paradigm, we proposed a geometry-centric self-supervised learning scheme, GC-MAE, which jointly reconstructs present 3D structure and forecasts its future evolution, enabling the vision encoder to capture both static geometry and physical dynamics. Furthermore, we developed a layer-wise temporal action modeling strategy that leverages intermediate and deep LLM representations to produce temporally coherent action sequences. Extensive experiments across simulation and real-world benchmarks demonstrate that Lift3D-VLA significantly outperforms prior VLA methods, achieving strong performance in multi-task, long-horizon, and dual-arm manipulation scenarios. Our method also exhibits robust generalization under diverse out-of-domain conditions, including unseen objects, lighting variations, and background clutter. These results highlight the effectiveness of combining explicit 3D reasoning with temporally aware action generation in dynamic environments. Despite these advances, several challenges remain. In particular, perception under transparent or reflective objects remains limited due to the inherent constraints of depth sensing. Future work will explore integrating depth completion and multi-view fusion to improve perceptual robustness, as well as developing closed-loop control mechanisms for finer-grained interaction in contact-rich settings.

{
    \small
    \bibliographystyle{IEEEtran}
    \bibliography{main}

@String(CVPR= {Proc. IEEE/CVF Conf. Comput. Vis. Pattern Recog.})

@String(ICCV= {Proc. Int. Conf. Comput. Vis.})

@String(ECCV= {Proc. Eur. Conf. Comput. Vis.})

@String(NIPS= {Proc. Adv. Neural Inform. Process. Syst.})

@String(ICLR = {Proc. Int. Conf. Learn. Represent.})

@String(ICML = {Proc. Int. Conf. Mach. Learn.})

@String(CORL = {Proc. Conf. Robot Learn.})

@String(RSS  = {Proc. Robot.: Sci. Syst.})

@String(RAL  = {IEEE Robot. Automat. Lett.})

@String(ICRA = {Proc. IEEE Int. Conf. Robot. Automat.})

@String(IROS = {Proc. IEEE/RSJ Int. Conf. Intell. Robots Syst.})

@String(TRO  = {IEEE Trans. Robot.})

@String(TMLR = {Trans. Mach. Learn. Res.})

@article{2024_9_05_OpenVLA,
  title={OpenVLA: An Open-Source Vision-Language-Action Model},
  author={Kim, Moo Jin and Pertsch, Karl and Karamcheti, Siddharth and Xiao, Ted and Balakrishna, Ashwin and Nair, Suraj and Rafailov, Rafael and Foster, Ethan and Lam, Grace and Sanketi, Pannag and others},
  journal={arXiv preprint arXiv:2406.09246},
  year={2024}
}

@article{2025_3_13_HybridVLA,
  title={HybridVLA: Collaborative Diffusion and Autoregression in a Unified Vision-Language-Action Model},
  author={Liu, Jiaming and Chen, Hao and An, Pengju and Liu, Zhuoyang and Zhang, Renrui and Gu, Chenyang and Li, Xiaoqi and Guo, Ziyu and Chen, Sixiang and Liu, Mengzhen and others},
  journal=ICLR,
  year={2025}
}

@article{2025_5_1_RDT_1B,
  title={Rdt-1b: a diffusion foundation model for bimanual manipulation},
  author={Liu, Songming and Wu, Lingxuan and Li, Bangguo and Tan, Hengkai and Chen, Huayu and Wang, Zhengyi and Xu, Ke and Su, Hang and Zhu, Jun},
  journal={arXiv preprint arXiv:2410.07864},
  year={2024}
}

@article{2024_10_31_pi0,
  title={pi0: A Vision-Language-Action Flow Model for General Robot Control},
  author={Black, Kevin and Brown, Noah and Driess, Danny and Esmail, Adnan and Equi, Michael and Finn, Chelsea and others},
  journal={arXiv preprint arXiv:2410.24164},
  year={2024}
}

@article{2025_3_18_GR00T,
  title={Gr00t n1: An open foundation model for generalist humanoid robots},
  author={Bjorck, Johan and Casta{\~n}eda, Fernando and Cherniadev, Nikita and Da, Xingye and Ding, Runyu and Fan, Linxi and Fang, Yu and Fox, Dieter and Hu, Fengyuan and Huang, Spencer and others},
  journal={arXiv preprint arXiv:2503.14734},
  year={2025}
}

@misc{2025_4_22_pi0_5,
      title={$\pi_{0.5}$: a Vision-Language-Action Model with Open-World Generalization}, 
      author={Physical Intelligence and Kevin Black and Noah Brown and James Darpinian and Karan Dhabalia and Danny Driess and Adnan Esmail and Michael Equi and Chelsea Finn and others},
      year={2025},
      eprint={2504.16054},
      archivePrefix={arXiv},
      primaryClass={cs.LG},
      url={https://arxiv.org/abs/2504.16054}, 
}

@inproceedings{goyal2023rvt,
  title={{RVT}: Robotic view transformer for {3D} object manipulation},
  author={Goyal, Ankit and Xu, Jie and Guo, Yijie and Blukis, Valts and Chao, Yu-Wei and Fox, Dieter},
  booktitle=CORL,
  pages={694--710},
  year={2023},
}

@inproceedings{jia2025lift3d,
  title={{Lift3D} Policy: Lifting {2D} Foundation Models for Robust {3D} Robotic Manipulation},
  author={Jia, Yueru and Liu, Jiaming and Chen, Sixiang and et al.},
  booktitle=CVPR,
  pages={17347--17358},
  year={2025}
}

@inproceedings{2025_8_08_3DS_VLA,
  title={3ds-vla: A 3d spatial-aware vision language action model for robust multi-task manipulation},
  author={Li, Xiaoqi and Heng, Liang and Liu, Jiaming and Shen, Yan and Gu, Chenyang and Liu, Zhuoyang and Chen, Hao and Han, Nuowei and Zhang, Renrui and Tang, Hao and others},
  booktitle=CORL,
  year={2025}
}

@inproceedings{2022_11_11_PerAct_ERCEIVER_ACTOR,
  title     = {Perceiver-Actor: A Multi-Task Transformer for Robotic Manipulation}, 
  author    = {Shridhar, Mohit and Manuelli, Lucas and Fox, Dieter},
  booktitle = CoRL,
  year      = {2022},
}

@inproceedings{gervet2023act3d,
  title={Act3d: 3d feature field transformers for multi-task robotic manipulation},
  author={Gervet, Theophile and Xian, Zhou and Gkanatsios, Nikolaos and Fragkiadaki, Katerina},
  booktitle=CORL,
  year={2023}
}

@inproceedings{2025_3_14_vggt,
  title={Vggt: Visual geometry grounded transformer},
  author={Wang, Jianyuan and Chen, Minghao and Karaev, Nikita and Vedaldi, Andrea and Rupprecht, Christian and Novotny, David},
  booktitle=CVPR,
  pages={5294--5306},
  year={2025}
}

@inproceedings{2021_2_26_CLIP,
  title={Learning transferable visual models from natural language supervision},
  author={Radford, Alec and Kim, Jong Wook and Hallacy, Chris and Ramesh, Aditya and Goh, Gabriel and Agarwal, Sandhini and Sastry, Girish and Askell, Amanda and Mishkin, Pamela and Clark, Jack and others},
  booktitle=ICML,
  pages={8748--8763},
  year={2021},
  organization={PMLR}
}

@article{2023_4_14_DINOv2,
  title={{DINO}v2: Learning Robust Visual Features without Supervision},
  author={Oquab, Maxime and Darcet, Timoth{\'e}e and Moutakanni, Th{\'e}o and et al.},
  journal=TMLR,
  year={2024}
}

@inproceedings{he2022masked,
  title={Masked autoencoders are scalable vision learners},
  author={He, Kaiming and Chen, Xinlei and Xie, Saining and Li, Yanghao and Doll{\'a}r, Piotr and Girshick, Ross},
  booktitle=CVPR,
  pages={16000--16009},
  year={2022}
}

@inproceedings{2020_7_21_PointContrast,
  title={Pointcontrast: Unsupervised pre-training for 3d point cloud understanding},
  author={Xie, Saining and Gu, Jiatao and Guo, Demi and Qi, Charles R and Guibas, Leonidas and Litany, Or},
  booktitle=ECCV,
  pages={574--591},
  year={2020},
  organization={Springer}
}

@article{li2026pointvla,
  title={Pointvla: Injecting the 3d world into vision-language-action models},
  author={Li, Chengmeng and Wen, Junjie and Peng, Yaxin and Peng, Yan and Zhu, Yichen},
  journal=RAL,
  volume={11},
  number={3},
  pages={2506--2513},
  year={2026},
  publisher={IEEE}
}

@article{2022_3_13_PointMAE,
  title={Masked autoencoders for 3d point cloud self-supervised learning},
  author={Pang, Yatian and Tay, Eng Hock Francis and Yuan, Li and Chen, Zhenghua},
  journal={World Scientific Annual Review of Artificial Intelligence},
  volume={1},
  pages={2440001},
  year={2023},
  publisher={World Scientific}
}

@article{2022_3_11_MVP, 
    title={Masked Visual Pre-training for Motor Control}, 
    author={Tete Xiao and Ilija Radosavovic and Trevor Darrell and Jitendra Malik}, 
    journal={arXiv preprint arXiv:2203.06173}, 
    year={2022} 
}

@article{2022_11_18_R3M,
  title={R3m: A universal visual representation for robot manipulation},
  author={Nair, Suraj and Rajeswaran, Aravind and Kumar, Vikash and Finn, Chelsea and Gupta, Abhinav},
  journal={arXiv preprint arXiv:2203.12601},
  year={2022}
}

@article{2023_3_31_VC_1,
  title={Where are we in the search for an artificial visual cortex for embodied intelligence?},
  author={Majumdar, Arjun and Yadav, Karmesh and Arnaud, Sergio and Ma, Jason and Chen, Claire and Silwal, Sneha and Jain, Aryan and Berges, Vincent-Pierre and Wu, Tingfan and Vakil, Jay and others},
  journal=NIPS,
  volume={36},
  pages={655--677},
  year={2023}
}

@article{khazatsky2024droid,
  title={Droid: A large-scale in-the-wild robot manipulation dataset},
  author={Khazatsky, Alexander and Pertsch, Karl and Nair, Suraj and Balakrishna, Ashwin and Dasari, Sudeep and Karamcheti, Siddharth and Nasiriany, Soroush and Srirama, Mohan Kumar and Chen, Lawrence Yunliang and Ellis, Kirsty and others},
  journal=RSS,
  year={2024}
}

@inproceedings{2024_2_12_Prismatic_VLMs,
  title={Prismatic vlms: Investigating the design space of visually-conditioned language models},
  author={Karamcheti, Siddharth and Nair, Suraj and Balakrishna, Ashwin and Liang, Percy and Kollar, Thomas and Sadigh, Dorsa},
  booktitle=ICML,
  year={2024}
}

@article{2017_6_PointNet_plus_plus,
  title={Pointnet++: Deep hierarchical feature learning on point sets in a metric space},
  author={Qi, Charles Ruizhongtai and Yi, Li and Su, Hao and Guibas, Leonidas J},
  journal=NIPS,
  volume={30},
  year={2017}
}

@misc{karamcheti2023languagedrivenrepresentationlearningrobotics,
      title={Language-Driven Representation Learning for Robotics}, 
      author={Siddharth Karamcheti and Suraj Nair and Annie S. Chen and Thomas Kollar and Chelsea Finn and Dorsa Sadigh and Percy Liang},
      year={2023},
      eprint={2302.12766},
      archivePrefix={arXiv},
      primaryClass={cs.RO},
      url={https://arxiv.org/abs/2302.12766}, 
}

@inproceedings{2021_6_17_LoRA,
  title={Lora: Low-rank adaptation of large language models},
  author={Hu, Edward J and Shen, Yelong and Wallis, Phillip and Allen-Zhu, Zeyuan and Li, Yuanzhi and Wang, Shean and Wang, Lu and Chen, Weizhu and others},
  journal=ICLR,
  year={2022}
}

@article{2019_9_26_RLBench,
  title={Rlbench: The robot learning benchmark \& learning environment},
  author={James, Stephen and Ma, Zicong and Arrojo, David Rovick and Davison, Andrew J},
  journal=RAL,
  volume={5},
  number={2},
  pages={3019--3026},
  year={2020},
  publisher={IEEE}
}

@inproceedings{2019_10_24_Meta_World,
  title={Meta-world: A benchmark and evaluation for multi-task and meta reinforcement learning},
  author={Yu, Tianhe and Quillen, Deirdre and He, Zhanpeng and Julian, Ryan and Hausman, Karol and Finn, Chelsea and Levine, Sergey},
  booktitle=CORL,
  pages={1094--1100},
  year={2020},
  organization={PMLR}
}

@inproceedings{qi2017pointnet,
  title={Pointnet: Deep learning on point sets for 3d classification and segmentation},
  author={Qi, Charles R and Su, Hao and Mo, Kaichun and Guibas, Leonidas J},
  booktitle=CVPR,
  pages={652--660},
  year={2017}
}

@article{2022_6_09_PointNeXt,
  title={Pointnext: Revisiting pointnet++ with improved training and scaling strategies},
  author={Qian, Guocheng and Li, Yuchen and Peng, Houwen and Mai, Jinjie and Hammoud, Hasan and Elhoseiny, Mohamed and Ghanem, Bernard},
  journal=NIPS,
  volume={35},
  pages={23192--23204},
  year={2022}
}

@inproceedings{ze20243d,
	title={3D Diffusion Policy: Generalizable Visuomotor Policy Learning via Simple 3D Representations},
	author={Yanjie Ze and Gu Zhang and Kangning Zhang and Chenyuan Hu and Muhan Wang and Huazhe Xu},
	booktitle=RSS,
	year={2024}
}

@article{zhu2024spa,
  title={Spa: 3d spatial-awareness enables effective embodied representation},
  author={Zhu, Haoyi and Yang, Honghui and Wang, Yating and Yang, Jiange and Wang, Limin and He, Tong},
  journal={arXiv preprint arXiv:2410.08208},
  year={2024}
}

@inproceedings{shridhar2023perceiver,
  title={Perceiver-actor: A multi-task transformer for robotic manipulation},
  author={Shridhar, Mohit and Manuelli, Lucas and Fox, Dieter},
  booktitle=CORL,
  pages={785--799},
  year={2023},
}

@misc{open_x_embodiment_rt_x_2023,
title={Open {X-E}mbodiment: Robotic Learning Datasets and {RT-X} Models},
author = {{Open X-Embodiment Collaboration} and Abhishek Padalkar and Acorn Pooley and others},
howpublished  = {\url{https://arxiv.org/abs/2310.08864}},
year = {2023},
}

@inproceedings{shridhar2022cliport,
  title={Cliport: What and where pathways for robotic manipulation},
  author={Shridhar, Mohit and Manuelli, Lucas and Fox, Dieter},
  booktitle=CORL,
  pages={894--906},
  year={2022},
  organization={PMLR}
}

@article{touvron2023llama2,
  title={Llama 2: Open foundation and fine-tuned chat models},
  author={Touvron, Hugo and Martin, Louis and Stone, Kevin and Albert, Peter and Almahairi, Amjad and Babaei, Yasmine and Bashlykov, Nikolay and Batra, Soumya and Bhargava, Prajjwal and Bhosale, Shruti and others},
  journal={arXiv preprint arXiv:2307.09288},
  year={2023}
}

@inproceedings{seo2023multi,
  title={Multi-view masked world models for visual robotic manipulation},
  author={Seo, Younggyo and Kim, Junsu and James, Stephen and Lee, Kimin and Shin, Jinwoo and Abbeel, Pieter},
  booktitle=ICML,
  pages={30613--30632},
  year={2023},
  organization={PMLR}
}

@inproceedings{qian20253d,
  title={3d-mvp: 3d multiview pretraining for manipulation},
  author={Qian, Shengyi and Mo, Kaichun and Blukis, Valts and Fouhey, David F and Fox, Dieter and Goyal, Ankit},
  booktitle=CVPR,
  pages={22530--22539},
  year={2025}
}

@article{yang2026hyperbolic,
  title={Hyperbolic Multiview Pretraining for Robotic Manipulation},
  author={Yang, Jin and Wei, Ping and Chen, Yixin},
  journal={arXiv preprint arXiv:2603.04848},
  year={2026}
}

@article{yang2026robo3r,
  title={Robo3R: Enhancing Robotic Manipulation with Accurate Feed-Forward 3D Reconstruction},
  author={Yang, Sizhe and Xu, Linning and Li, Hao and Mu, Juncheng and Zeng, Jia and Lin, Dahua and Pang, Jiangmiao},
  journal={arXiv preprint arXiv:2602.10101},
  year={2026}
}

@article{zhang2025canonical,
  title={Canonical Policy: Learning Canonical 3D Representation for SE (3)-Equivariant Policy},
  author={Zhang, Zhiyuan and Xu, Zhengtong and Lakamsani, Jai Nanda and She, Yu},
  journal={arXiv preprint arXiv:2505.18474},
  year={2025}
}

@article{cui2025cl3r,
  title={{CL3R}: 3D Reconstruction and Contrastive Learning for Enhanced Robotic Manipulation Representations},
  author={Cui, Wenbo and Zhao, Chengyang and Chen, Yuhui and et al.},
  journal={arXiv preprint arXiv:2507.08262},
  year={2025}
}

@article{ho2020denoising,
  title={Denoising diffusion probabilistic models},
  author={Ho, Jonathan and Jain, Ajay and Abbeel, Pieter},
  journal=NIPS,
  volume={33},
  pages={6840--6851},
  year={2020}
}

@inproceedings{zitkovich2023rt,
  title={Rt-2: Vision-language-action models transfer web knowledge to robotic control},
  author={Zitkovich, Brianna and Yu, Tianhe and Xu, Sichun and Xu, Peng and Xiao, Ted and Xia, Fei and Wu, Jialin and Wohlhart, Paul and Welker, Stefan and Wahid, Ayzaan and others},
  booktitle=CORL,
  year={2023}
}

@inproceedings{ddim2021,
  title={Denoising Diffusion Implicit Models},
  author={Jiaming Song, Chenlin Meng, Stefano Ermon},
  booktitle=ICLR,
  year={2021}
}

@article{caron2020unsupervised,
  title={Unsupervised learning of visual features by contrasting cluster assignments},
  author={Caron, Mathilde and Misra, Ishan and Mairal, Julien and Goyal, Priya and Bojanowski, Piotr and Joulin, Armand},
  journal=NIPS,
  volume={33},
  pages={9912--9924},
  year={2020}
}

@article{fu2024mobile,
  title={Mobile ALOHA: Learning Bimanual Mobile Manipulation with Low-Cost Whole-Body Teleoperation},
  author={Fu, Zipeng and Zhao, Tony Z and Finn, Chelsea},
  journal={arXiv preprint arXiv:2401.02117},
  year={2024}
}

@misc{li2025vipvisioninstructedpretraining,
      title={VIP: Vision Instructed Pre-training for Robotic Manipulation}, 
      author={Zhuoling Li and Liangliang Ren and Jinrong Yang and Yong Zhao and Xiaoyang Wu and Zhenhua Xu and Xiang Bai and Hengshuang Zhao},
      year={2025},
      eprint={2410.07169},
      archivePrefix={arXiv},
      primaryClass={cs.RO},
      url={https://arxiv.org/abs/2410.07169}, 
}

@article{zhu2024point,
  title={Point cloud matters: Rethinking the impact of different observation spaces on robot learning},
  author={Zhu, Haoyi and Wang, Yating and Huang, Di and Ye, Weicai and Ouyang, Wanli and He, Tong},
  journal=NIPS,
  volume={37},
  pages={77799--77830},
  year={2024}
}

@inproceedings{zhai2023siglip,
  author = {Xiaohua Zhai and Basil Mustafa and Alexander Kolesnikov and Lucas Beyer},
  booktitle = ICCV,
  title = {Sigmoid Loss for Language Image Pre-Training},
  year = {2023},
}

@article{james2022coarse,
  title={Coarse-to-fine q-attention with learned path ranking},
  author={James, Stephen and Abbeel, Pieter},
  journal={arXiv preprint arXiv:2204.01571},
  year={2022}
}

@inproceedings{xian2023chaineddiffuser,
  title={Chaineddiffuser: Unifying trajectory diffusion and keypose prediction for robotic manipulation},
  author={Xian, Zhou and Gkanatsios, Nikolaos and Gervet, Theophile and Ke, Tsung-Wei and Fragkiadaki, Katerina},
  booktitle=CORL,
  year={2023}
}

@inproceedings{liurdt,
  title={RDT-1B: a Diffusion Foundation Model for Bimanual Manipulation},
  author={Liu, Songming and Wu, Lingxuan and Li, Bangguo and Tan, Hengkai and Chen, Huayu and Wang, Zhengyi and Xu, Ke and others},
  booktitle=ICLR
}

@inproceedings{o2024open,
  title={Open X-Embodiment: Robotic Learning Datasets and RT-X Models},
  author={O'Neill, Abby and Rehman, Abdul and Maddukuri, Abhiram and et al.},
  booktitle=ICRA,
  pages={6892--6903},
  year={2024}
}

@article{fang2023anygrasp,
  title={Anygrasp: Robust and efficient grasp perception in spatial and temporal domains},
  author={Fang, Hao-Shu and Wang, Chenxi and Fang, Hongjie and Gou, Minghao and Liu, Jirong and Yan, Hengxu and Liu, Wenhai and Xie, Yichen and Lu, Cewu},
  journal=TRO,
  year={2023},
  publisher={IEEE}
}

@article{qu2025spatialvla,
  title={Spatialvla: Exploring spatial representations for visual-language-action model},
  author={Qu, Delin and Song, Haoming and Chen, Qizhi and Yao, Yuanqi and Ye, Xinyi and Ding, Yan and Wang, Zhigang and Gu, JiaYuan and Zhao, Bin and Wang, Dong and others},
  journal={arXiv preprint arXiv:2501.15830},
  year={2025}
}

@inproceedings{zhao2025cot,
  title={Cot-vla: Visual chain-of-thought reasoning for vision-language-action models},
  author={Zhao, Qingqing and Lu, Yao and Kim, Moo Jin and Fu, Zipeng and Zhang, Zhuoyang and Wu, Yecheng and Li, Zhaoshuo and Ma, Qianli and Han, Song and Finn, Chelsea and others},
  booktitle=CVPR,
  pages={1702--1713},
  year={2025}
}

@article{liu2022frame,
  title={Frame mining: a free lunch for learning robotic manipulation from 3d point clouds},
  author={Liu, Minghua and Li, Xuanlin and Ling, Zhan and Li, Yangyan and Su, Hao},
  journal=CoRL,
  year={2022}
}

@article{chen2023polarnet,
  title={Polarnet: 3d point clouds for language-guided robotic manipulation},
  author={Chen, Shizhe and Garcia, Ricardo and Schmid, Cordelia and Laptev, Ivan},
  journal=CoRL,
  year={2023}
}

@article{yue2024deer,
  title={DeeR-VLA: Dynamic Inference of Multimodal Large Language Models for Efficient Robot Execution},
  author={Yue, Yang and Wang, Yulin and Kang, Bingyi and Han, Yizeng and Wang, Shenzhi and Song, Shiji and Feng, Jiashi and Huang, Gao},
  journal={arXiv preprint arXiv:2411.02359},
  year={2024}
}

@article{ke20243d,
  title={3d diffuser actor: Policy diffusion with 3d scene representations},
  author={Ke, Tsung-Wei and Gkanatsios, Nikolaos and Fragkiadaki, Katerina},
  journal={arXiv preprint arXiv:2402.10885},
  year={2024}
}

@article{wang2024vihe,
  title={Vihe: Virtual in-hand eye transformer for 3d robotic manipulation},
  author={Wang, Weiyao and Lei, Yutian and Jin, Shiyu and Hager, Gregory D and Zhang, Liangjun},
  journal={arXiv preprint arXiv:2403.11461},
  year={2024}
}

@article{wen2024diffusion,
  title={Diffusion-VLA: Scaling Robot Foundation Models via Unified Diffusion and Autoregression},
  author={Wen, Junjie and Zhu, Minjie and Zhu, Yichen and Tang, Zhibin and Li, Jinming and Zhou, Zhongyi and Li, Chengmeng and Liu, Xiaoyu and Peng, Yaxin and Shen, Chaomin and others},
  journal={arXiv preprint arXiv:2412.03293},
  year={2024}
}

@inproceedings{wu2024robomind,
  title={Robomind: Benchmark on multi-embodiment intelligence normative data for robot manipulation},
  author={Wu, Kun and Hou, Chengkai and Liu, Jiaming and Che, Zhengping and Ju, Xiaozhu and others},
	booktitle=RSS, 
  year={2025},
  publisher={Robotics: Science and Systems Foundation}, 
}

@article{chen2025fast,
  title={Fast-in-Slow: A Dual-System Foundation Model Unifying Fast Manipulation within Slow Reasoning},
  author={Chen, Hao and Liu, Jiaming and Gu, Chenyang and Liu, Zhuoyang and Zhang, Renrui and Li, Xiaoqi and He, Xiao and Guo, Yandong and Fu, Chi-Wing and Zhang, Shanghang and others},
  journal={arXiv preprint arXiv:2506.01953},
  year={2025}
}

@article{bjorck2025gr00t,
  title={Gr00t n1: An open foundation model for generalist humanoid robots},
  author={Bjorck, Johan and Casta{\~n}eda, Fernando and Cherniadev, Nikita and Da, Xingye and Ding, Runyu and Fan, Linxi and Fang, Yu and Fox, Dieter and Hu, Fengyuan and Huang, Spencer and others},
  journal={arXiv preprint arXiv:2503.14734},
  year={2025}
}

@article{zhang2025upvlaunifiedunderstandingprediction,
  title={Up-vla: A unified understanding and prediction model for embodied agent},
  author={Zhang, Jianke and Guo, Yanjiang and Hu, Yucheng and Chen, Xiaoyu and Zhu, Xiang and Chen, Jianyu},
  booktitle=ICML,
  year={2025}
}

@article{liu2025mla,
  title={MLA: A Multisensory Language-Action Model for Multimodal Understanding and Forecasting in Robotic Manipulation},
  author={Liu, Zhuoyang and Liu, Jiaming and Xu, Jiadong and Han, Nuowei and Gu, Chenyang and Chen, Hao and Zhou, Kaichen and Zhang, Renrui and Hsieh, Kai Chin and Wu, Kun and others},
  journal={arXiv preprint arXiv:2509.26642},
  year={2025}
}

@article{huang2025thinkact,
  title={Thinkact: Vision-language-action reasoning via reinforced visual latent planning},
  author={Huang, Chi-Pin and Wu, Yueh-Hua and Chen, Min-Hung and Wang, Yu-Chiang Frank and Yang, Fu-En},
  journal={arXiv preprint arXiv:2507.16815},
  year={2025}
}

@article{gu2025manualvla,
  title={ManualVLA: A Unified VLA Model for Chain-of-Thought Manual Generation and Robotic Manipulation},
  author={Gu, Chenyang and Liu, Jiaming and Chen, Hao and Huang, Runzhong and Wuwu, Qingpo and Liu, Zhuoyang and Li, Xiaoqi and Li, Ying and Zhang, Renrui and Jia, Peng and others},
  journal=CVPR,
  year={2025}
}

@article{hou2025robomind,
  title={RoboMIND 2.0: A Multimodal, Bimanual Mobile Manipulation Dataset for Generalizable Embodied Intelligence},
  author={Hou, Chengkai and Wu, Kun and Liu, Jiaming and Che, Zhengping and Wu, Di and Liao, Fei and Li, Guangrun and He, Jingyang and Feng, Qiuxuan and Jin, Zhao and others},
  journal={arXiv preprint arXiv:2512.24653},
  year={2025}
}

@inproceedings{eisner2022flowbot3d,
  title={FlowBot3D: Learning 3D Articulation Flow to Manipulate Articulated Objects},
  author={Eisner*, Ben and Zhang*, Harry and Held,David},
  booktitle=RSS,
  year={2022}
}

@INPROCEEDINGS{wang2024rise,
  author={Wang, Chenxi and Fang, Hongjie and Fang, Hao-Shu and Lu, Cewu},
  booktitle=IROS, 
  title={RISE: 3D Perception Makes Real-World Robot Imitation Simple and Effective}, 
  year={2024},
}

@article{zhang2024leveraging,
  title={Leveraging Locality to Boost Sample Efficiency in Robotic Manipulation},
  author={Zhang, Tong and Hu, Yingdong and You, Jiacheng and Gao, Yang},
  journal=CoRL,
  year={2024}
}

@inproceedings{zhang2024sam,
  title={SAM-E: Leveraging Visual Foundation Model with Sequence Imitation for Embodied Manipulation},
  author={Zhang, Junjie and Bai, Chenjia and He, Haoran and Wang, Zhigang and Zhao, Bin and Li, Xiu and Li, Xuelong},
  booktitle=ICML,
  pages={58579--58598},
  year={2024},
  organization={PMLR}
}

@inproceedings{chenac,
  title={AC-DiT: Adaptive Coordination Diffusion Transformer for Mobile Manipulation},
  author={Chen, Sixiang and Liu, Jiaming and Qian, Siyuan and Jiang, Han and Liu, Zhuoyang and Gu, Chenyang and Li, Xiaoqi and Hou, Chengkai and Wang, Pengwei and Wang, Zhongyuan and others},
  booktitle=NIPS
}

@article{bi2025motus,
  title={Motus: A unified latent action world model},
  author={Bi, Hongzhe and Tan, Hengkai and Xie, Shenghao and Wang, Zeyuan and Huang, Shuhe and Liu, Haitian and Zhao, Ruowen and Feng, Yao and Xiang, Chendong and Rong, Yinze and others},
  journal={arXiv preprint arXiv:2512.13030},
  year={2025}
}

@article{liu2026last,
  title={{LaST$_0$}: Latent Spatio-Temporal Chain-of-Thought for Robotic Vision-Language-Action Model},
  author={Liu, Zhuoyang and Liu, Jiaming and Chen, Hao and et al.},
  journal={arXiv preprint arXiv:2601.05248},
  year={2026}
}

@article{kim2026cosmos,
  title={Cosmos policy: Fine-tuning video models for visuomotor control and planning},
  author={Kim, Moo Jin and Gao, Yihuai and Lin, Tsung-Yi and Lin, Yen-Chen and Ge, Yunhao and Lam, Grace and Liang, Percy and Song, Shuran and Liu, Ming-Yu and Finn, Chelsea and others},
  journal={arXiv preprint arXiv:2601.16163},
  year={2026}
}

@article{sucan2012open,
  title={The open motion planning library},
  author={Sucan, Ioan A and Moll, Mark and Kavraki, Lydia E},
  journal={IEEE Robotics \& Automation Magazine},
  volume={19},
  number={4},
  pages={72--82},
  year={2012},
  publisher={IEEE}
}

@article{chen2021decision,
  title={Decision transformer: Reinforcement learning via sequence modeling},
  author={Chen, Lili and Lu, Kevin and Rajeswaran, Aravind and Lee, Kimin and Grover, Aditya and Laskin, Misha and Abbeel, Pieter and Srinivas, Aravind and Mordatch, Igor},
  journal={Advances in neural information processing systems},
  volume={34},
  pages={15084--15097},
  year={2021}
}

@article{wen2023large,
  title={Large sequence models for sequential decision-making: a survey},
  author={Wen, Muning and Lin, Runji and Wang, Hanjing and Yang, Yaodong and Wen, Ying and Mai, Luo and Wang, Jun and Zhang, Haifeng and Zhang, Weinan},
  journal={Frontiers of Computer Science},
  volume={17},
  number={6},
  pages={176349},
  year={2023},
  publisher={Springer}
}

@article{chen2026mv,
  title={MV-WAM: Manifold-Aware World Action Model with Value Augmentation},
  author={Chen, Jintao and Jia, Peidong and Wuwu, Qingpo and Liu, Jiaming and Du, Mengfei and Fan, Chun-Kai and Chi, Xiaowei and Chen, Hao and Bai, Chengyu and Qian, Zezhong and others},
  journal={arXiv preprint arXiv:2606.21088},
  year={2026}
}

@article{liu2026last_hd,
  title={LaST-HD: Learning Latent Physical Reasoning from Scalable Human Data for Robot Manipulation},
  author={Liu, Jiaming and Wang, Yinxi and Gu, Chenyang and Qian, Siyuan and Mi, Xiangju and Chen, Hao and Chen, Jiawei and Wuwu, Qingpo and Li, Xiaoqi and Han, Nuowei and others},
  journal={arXiv preprint arXiv:2606.23685},
  year={2026}
}
}

\end{document}